\documentclass[final]{cvpr}

\usepackage{epsfig}
\usepackage{graphicx}
\usepackage{amsmath}
\usepackage{amssymb}

\usepackage{threeparttable}
\usepackage[vlined, ruled, linesnumbered]{algorithm2e}
\usepackage{multirow}
\usepackage{booktabs}
\usepackage{bbm}
\usepackage{pifont}
\usepackage{subfig}
\usepackage{graphicx}
\usepackage{amsmath}
\usepackage{enumitem}

\usepackage{eucal,nicefrac}

\usepackage{bm}
\usepackage{mathrsfs}

\let\mathscr=\mathcal

\makeatletter
\renewcommand{\paragraph}{%
  \@startsection{paragraph}{4}%
  {\z@}{0.5ex \@plus 1ex \@minus .1ex}{-1em}%
  {\normalfont\normalsize\bfseries}%
}
\makeatother

\AtBeginDocument{%
 \abovedisplayskip=8pt plus 3pt minus 5pt
 \abovedisplayshortskip=0pt plus 3pt
 \belowdisplayskip=8pt plus 3pt minus 5pt
 \belowdisplayshortskip=5pt plus 3pt minus 4pt
}

\usepackage[pagebackref=true,breaklinks=true,colorlinks,bookmarks=false]{hyperref}

\usepackage[font=small,labelfont=bf,tableposition=top]{caption}

\begin{document}

\title{Learning to Recover 3D Scene Shape from a Single Image
}

\author{
Wei Yin$^\dag $,
Jianming Zhang$^\ddag$,
Oliver Wang$^\ddag$,
Simon Niklaus$^\ddag$, 
Long Mai$^\ddag$, 
Simon Chen$^\ddag$, 
Chunhua Shen$^\dag$\thanks{Correspondence should be addressed to C. Shen.}
\\[0.2cm]
$ ^\dag$ The University of Adelaide, Australia
~~~~~~~~~~~~~~
$ ^\ddag$ Adobe Research
}

\makeatletter
\let\@oldmaketitle\@maketitle%
\renewcommand{\@maketitle}{\@oldmaketitle%
 \centering
    \includegraphics[width=1\textwidth]{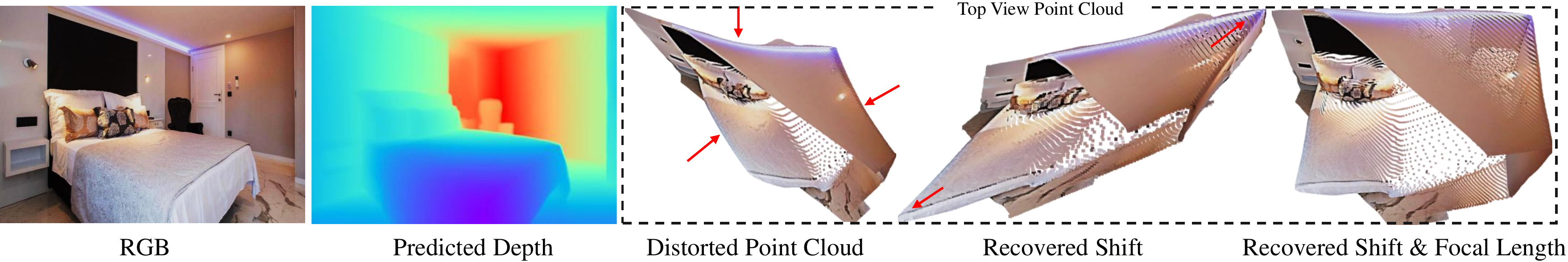}
     \captionof{figure}{3D scene structure distortion of projected point clouds. 
    While the predicted depth map is correct, the 3D scene shape of the point cloud suffers from noticeable distortions due to an unknown depth shift and focal length (third column). Our method recovers these parameters using 3D point cloud networks. With recovered depth shift, the walls and bed edges become straight, but the overall scene is stretched (fourth column). Finally, with recovered focal length, an accurate 3D scene can be reconstructed (fifth column).}
    \label{Fig: first page fig.}
    \bigskip}                   %
\makeatother

\maketitle

\begin{abstract}

Despite significant progress in monocular depth estimation in the wild, recent state-of-the-art methods cannot be used to recover accurate 3D scene shape due to an unknown depth shift induced by shift-invariant reconstruction losses used in mixed-data depth prediction training, and possible unknown camera focal length.
We investigate this problem in detail, and propose a two-stage framework that first predicts depth up to an unknown scale and shift from a single monocular image, and then use 3D point cloud encoders to predict the missing depth shift and focal length that allow us to recover a realistic 3D scene shape.
In addition, we propose an image-level normalized regression loss and a normal-based geometry loss to enhance depth prediction models trained on mixed datasets. 
We test our depth model on nine unseen datasets and 
achieve state-of-the-art performance
on zero-shot dataset generalization.
Code is available at:
{   
    \def\UrlFont{\rm\small\ttfamily}
\url{https://git.io/Depth}
}
%
%
%

\end{abstract}

\section{Introduction}
3D scene reconstruction is a fundamental task in computer vision. 
The established approach to address this task is SLAM or SfM~\cite{hartley2003multiple}, which reconstructs 3D 
scenes based on feature-point correspondence with consecutive frames or multiple views. 
In contrast, this work aims to achieve \textit{ 
dense 3D scene shape reconstruction from a single in-the-wild image}. 
Without multiple views available, we rely on monocular depth estimation. 
However, as shown in \figref{Fig: first page fig.}, existing monocular depth estimation methods~\cite{eigen2014depth, wang2020sdc, Yin2019enforcing} alone are unable to faithfully recover an accurate 3D point cloud.

Unlike multi-view reconstruction methods, monocular depth estimation requires leveraging high level scene priors, so data-driven approaches have become the \textit{de facto} solution to this problem~\cite{li2018megadepth, Ranftl2020, wang2019web, yin2020diversedepth}. 
Recent works have shown promising results by training deep neural networks on diverse in-the-wild data, \eg web stereo images and stereo videos~\cite{chen2016single, chen2020oasis, Ranftl2020, wang2019web,  xian2018monocular, xian2020structure, yin2020diversedepth}.
However, the diversity of the training data also poses challenges for the model training, as training data captured by different cameras can exhibit significantly different image priors for depth estimation~\cite{facil2019cam}. 
Moreover, web stereo images and videos can only provide depth supervision up to a scale and shift due to the unknown camera baselines and stereoscopic post processing~\cite{lang2010nonlinear}. 
As a result, state-of-the-art in-the-wild monocular depth models use various types of losses invariant to scale and shift in training. 
While an unknown scale in depth will not cause any shape distortion, as it scales the 3D scene uniformly, an unknown depth shift will (see Sec.~\ref{sec:3D_shape} and Fig.~\ref{Fig: first page fig.}). 
In addition, the camera focal length of a given image may not be accessible at test time, leading to more distortion of the 3D scene shape. 
This scene shape distortion is a critical problem for downstream tasks such as 3D view synthesis and 3D photography.

To address these challenges, we propose a novel monocular scene shape estimation framework that consists of a depth prediction module and a point cloud reconstruction module. 
The depth prediction module is a convolutional neural network trained on a mixture of existing datasets that predicts depth maps up to a scale and shift. 
The point cloud reconstruction module leverages point cloud encoder networks that predict shift and focal length adjustment factors from an initial guess of the scene point cloud reconstruction. 
A key observation 
that 
we make here is that,  \textit{when operating on point clouds derived from depth maps, and not on images themselves, we can train models to learn 3D scene shape priors using synthetic 3D data or data acquired by 3D laser scanning devices.
The domain gap is
significantly less of an issue for point clouds than that for images, 
}%
although these data sources are significantly less diverse than internet images. 

We 
empirically 
show that these point cloud encoders generalize well to unseen datasets.

Furthermore, to train a robust monocular depth prediction model on mixed data from multiple sources, we propose a simple but effective image-level normalized regression loss, and a pair-wise surface normal regression loss.
The former loss transforms the depth data to a canonical scale-shift-invariant space for more robust training, while the latter improves the geometry of our predicted depth maps. 
To summarize, our main contributions are: 
\begin{itemize}[noitemsep]
    \item A novel framework for in-the-wild monocular 3D scene shape estimation. To the best of our knowledge, this is the first fully data-driven method for this task, and the first method to leverage 3D point cloud neural networks for improving the structure of point clouds derived from depth maps. 
    \item An image-level normalized regression loss and a pair-wise surface normal regression loss for improving monocular depth estimation models trained on mixed multi-source datasets. 
\end{itemize}
Experiments show that our point cloud reconstruction module can recover accurate 3D shape from a single image, 
and that our depth prediction module achieves state-of-the-art results on zero-shot dataset transfer to $9$ unseen datasets.

\section{Related Work}

\paragraph{Monocular depth estimation in the wild.} 
This task has recently seen impressive progress~\cite{chen2016single, chen2019learning, chen2020oasis, li2018megadepth, wang2019web,  wang2020foresee, xian2018monocular, xian2020structure, yin2020diversedepth}. 
The key properties of such approaches are what data can be used for training, and what objective function makes sense for that data. 
When metric depth supervision is available, networks can be trained to directly regress these depths~\cite{eigen2014depth, liu2015learning, Yin2019enforcing}. 
However, obtaining metric ground truth depth for diverse datasets is challenging. 
As an alternative, Chen~\etal~\cite{chen2016single} collect diverse \emph{relative} depth annotations for internet images, while other approaches propose to scrape stereo images or videos from the internet~\cite{Ranftl2020, wang2019web,  xian2018monocular, xian2020structure, yin2020diversedepth}.
Such diverse data is important for generalizability, but as the metric depth is not available, direct depth regression losses cannot be used. 
Instead, these methods rely either on ranking losses which evaluate relative depth~\cite{chen2016single, xian2018monocular, xian2020structure} or scale and shift invariant losses~\cite{Ranftl2020, wang2019web} for supervision. 
The later methods produce especially robust depth predictions, but as the camera model is unknown and an unknown shift resides in the depth, the 3D shape cannot be reconstructed from the predicted depth maps. 
In this paper, we aim to reconstruct the 3D shape from a single image in the wild.

\paragraph{3D reconstruction from a single image.} 
A number of works have addressed reconstructing different types of objects from a single image~\cite{barron2014shape, wang2018pixel2mesh, wu2018learning}, such as humans~\cite{saito2019pifu, saito2020pifuhd}, cars, planes, tables, etc. 
The main challenge is how to best recover objects details, and how to represent them with limited memory. 
Pixel2Mesh~\cite{wang2018pixel2mesh} proposes to reconstruct the 3D shape from a single image and express it in a triangular mesh.
PIFu~\cite{saito2019pifu, saito2020pifuhd} proposes an memory-efficient implicit function to recover high-resolution surfaces, including unseen/occluded regions, of humans.
However, all these methods rely on learning priors specific to a certain object class or instance, typically from 3D supervision, and can therefore not work for full scene reconstruction.

On the other hand, several works have proposed reconstructing 3D scene structure from a single image.
Saxena~\etal~\cite{saxena2008make3d} assume that the whole scene can be segmented into several pieces, of which each one can be regarded as a small plane. 
They predict the orientation and the location of the planes and stitch them together to represent the scene. 
Other works propose to use image cues, such as shading~\cite{prados2005shape} and contour edges~\cite{karpenko2006smoothsketch} for scene reconstruction. 
However, these approaches use hand-designed priors and restrictive assumptions about the scene geometry. 
Our method is fully data driven, and can be applied to a wide range of scenes.

\begin{figure*}[t]
\centering
\includegraphics[width=1\textwidth]{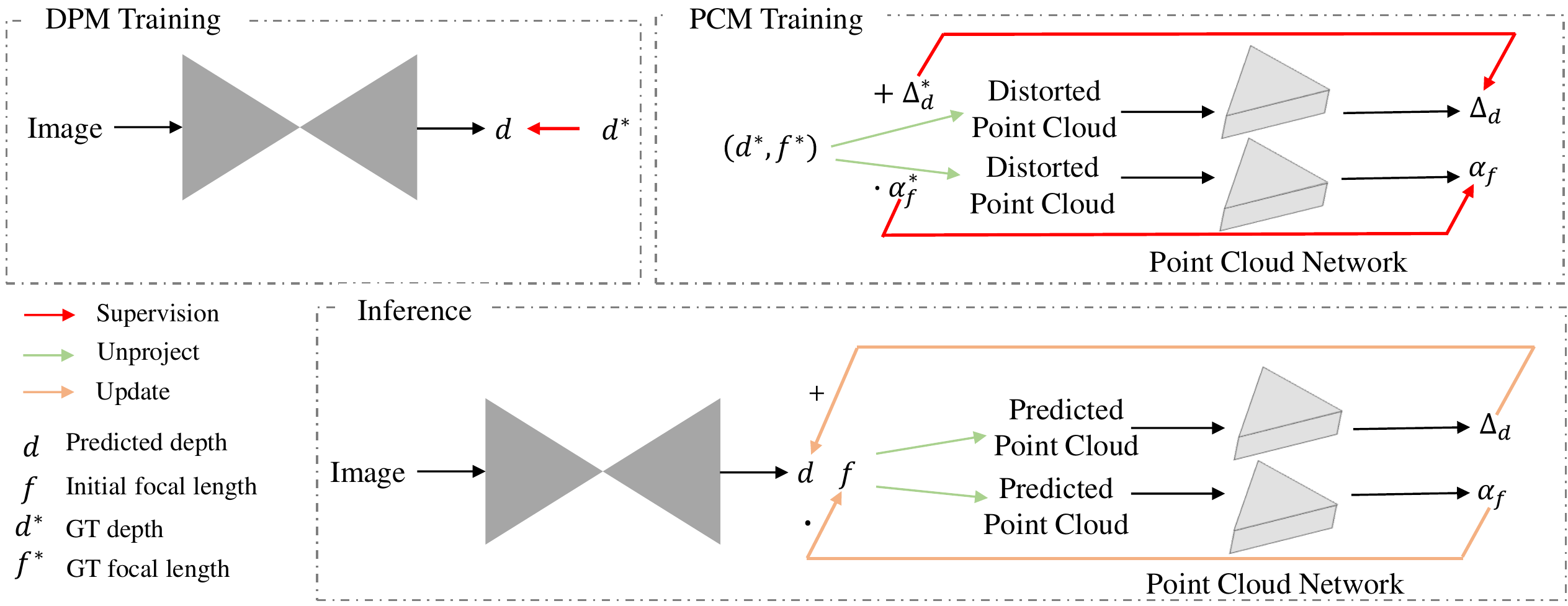}
\caption{\textbf{Method Pipeline.} During training, the depth prediction model (top left) and point cloud module (top right) are trained separately on different sources of data. During inference (bottom), the two networks are combined together to predict depth $d$ and from that, the depth shift $\Delta_{d}$ and focal length $f\cdot\alpha_{f}$ that together allow for an accurate scene shape reconstruction. Note that we employ point cloud networks to predict shift and focal length scaling factor separately. Please see the text for more details.}
\label{Fig: pipeline}
\vspace{-0.5em}
\end{figure*}

\paragraph{Camera intrinsic parameter estimation.} 
Recovering a camera's focal length is an important part of 3D scene understanding. 
Traditional methods utilize reference objects such as a planar calibration grids~\cite{zhang2000flexible} or vanishing points~\cite{deutscher2002automatic}, which can then be used to estimate a focal length. 
Other methods~\cite{hold2018perceptual, workman2015deepfocal} propose a data driven approach where a CNN recovers the focal length on in-the-wild data directly from an image.
In contrast, our point cloud module estimates the focal length directly in 3D, which we argue is an easier task than operating on natural images directly.

\section{Method}
Our two-stage single image 3D shape estimation pipeline is illustrated in Fig.~\ref{Fig: pipeline}. 
It is composed of a depth prediction module (DPM) and a point cloud module (PCM). 
The two modules are trained separately on different data sources, and are then combined together at inference time. 
The DPM takes an RGB image and outputs a depth map~\cite{yin2020diversedepth} with unknown scale and shift in relation to the true metric depth map. 
The PCM takes as input a distorted 3D point cloud, computed using a predicted depth map $d$ and an initial estimation of the focal length $f$, and outputs shift adjustments to the depth map and focal length to improve the geometry of the reconstructed 3D scene shape. 

\subsection{Point Cloud Module}\label{sec:3D_shape}

\begin{figure}[t]
\centering
\includegraphics[width=\linewidth]{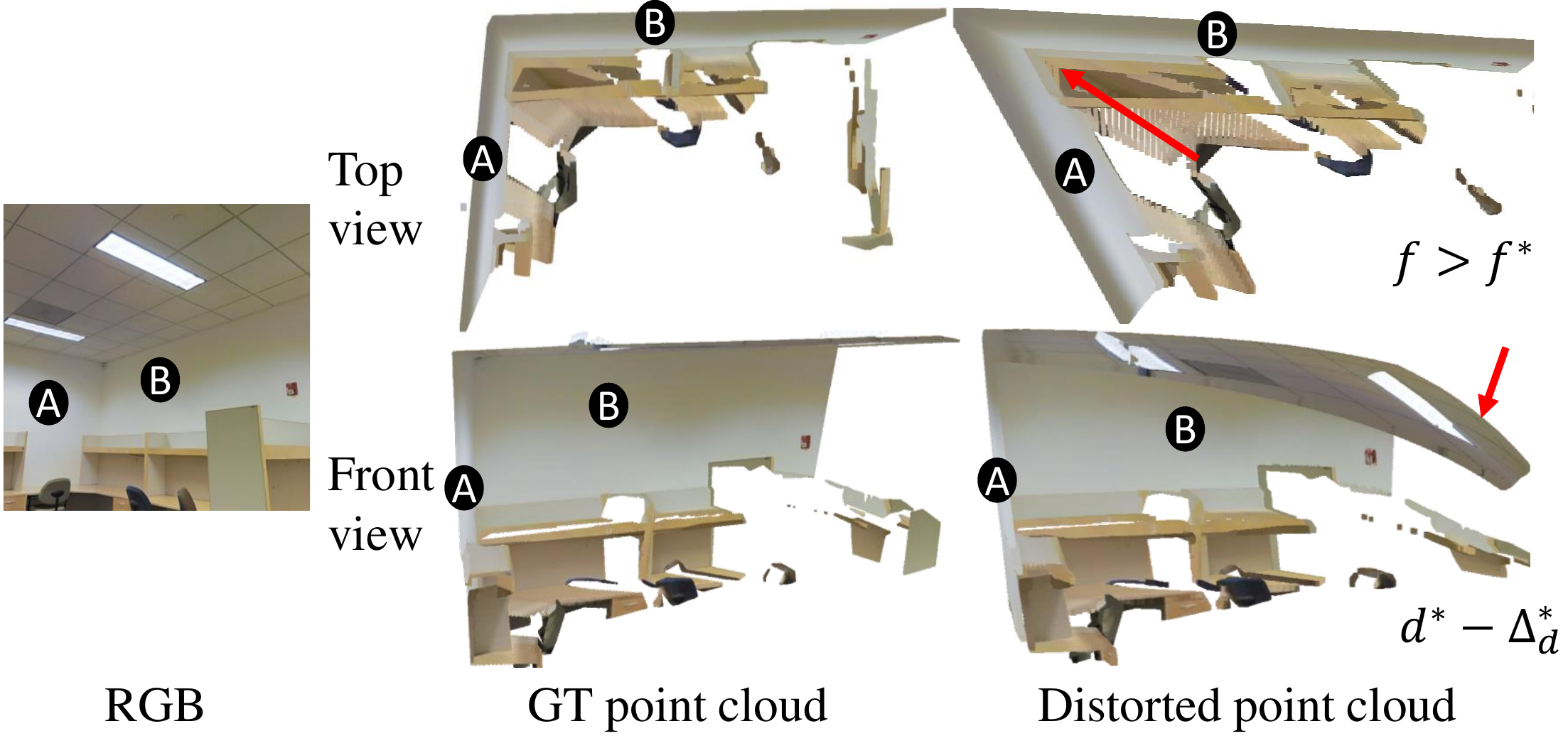}
\caption{Illustration of the distorted 3D shape caused by incorrect shift and focal length. A ground truth depth map is projected in 3D and visualized. When the focal length is incorrectly estimated ($f>f^{*}$), we observe significant structural distortion, e.g., see the angle between two walls A and B. Second row (front view): a shift ($d^{*}+\Delta_{d}$) also causes the shape distortion, see the roof. }
\label{Fig: focal length and shift affect shape. }
\vspace{-0.5em}
\end{figure}

We assume a pinhole camera model for the 3D point cloud reconstruction, which means that the unprojection from 2D coordinates and depth to 3D points is:
\begin{equation}
    \left\{\begin{matrix}x =&\hspace{-0.3cm} \frac{u-u_{0}}{f}d
    \\ y =&\hspace{-0.3cm} \frac{v-v_{0}}{f}d
    \\ z =&\hspace{-0.3cm} \hspace{0.75cm}d
    \end{matrix}\right.
\label{eq: 3D point cloud reconstruction}
\end{equation}
where $(u_{0}, v_{0})$ are the camera optical center, $f$ is the focal length, and $d$ is the depth. 
The focal length affects the point cloud shape as it scales $x$ and $y$ coordinates, but not $z$.
Similarly, a shift of $d$ will affect the $x$, $y$, and $z$ coordinates non-uniformly, which will result in shape distortions. 

For a human observer, these distortions are immediately recognizable when viewing the point cloud at an oblique angle (Fig.~\ref{Fig: focal length and shift affect shape. }), although they cannot be observed looking at a depth map alone. 
As a result, we propose to directly analyze the point cloud to determine the unknown shift and focal length parameters. 
We tried a number of network architectures that take unstructured 3D point clouds as input, and found that the recent PVCNN~\cite{liu2019pvcnn} performed well for this task, so we use it in all experiments here.
During training, a perturbed input point cloud with incorrect shift and focal length is synthesized by perturbing the known ground truth depth shift and focal length. 
The ground truth depth $d^{*}$ is transformed by a shift $\Delta^{*}_{d}$ drawn from $\mathcal{U}(-0.25, 0.8)$, and the ground truth focal length $f^*$ is transformed by a scale $\alpha^{*}_{f}$ drawn from $\mathcal{U}(0.6, 1.25)$ to keep the focal length positive and non-zero.

When recovering the depth shift, the perturbed 3D point cloud is $\mathcal{F}(u_{0}, v_{0}, f^{*}, d^{*} + \Delta^{*}_{d})$ is given as input to the shift point cloud network $\mathcal{N}_{d}(\cdot)$, trained with the objective: 
\begin{equation}
    L = \min_{\theta}\left | \mathcal{N}_{d}(\mathcal{F}(u_{0}, v_{0}, f^{*}, d^{*} + \Delta^{*}_{d}), \theta) - \Delta^{*}_{d} \right | 
\label{eq: regress shift}
\end{equation}
where $\theta$ are network weights and $f^{*}$ is the true focal length. 

Similarly, when recovering the focal length, the point cloud $\mathcal{F}(u_{0}, v_{0}, \alpha^*_f f^*, d^{*})$ is fed to the focal length point cloud network $\mathcal{N}_{f}(\cdot)$, trained with the objective:
\begin{equation}
    L = \min_{\theta}\left | \mathcal{N}_{f}(\mathcal{F}(u_{0}, v_{0}, \alpha^{*}_{f}f^{*}, d^{*}), \theta) - \alpha^{*}_{f} \right | 
\end{equation}

During inference, the ground truth depth is replaced with the predicted affine-invariant depth $d$, which is normalized to $[0, 1]$ prior to the 3D reconstruction. 
We use an initial guess of focal length $f$, giving us the reconstructed point cloud $\mathcal{F}(u_{0}, v_{0}, f, d)$, which is fed to $\mathcal{N}_{d}(\cdot)$ and $\mathcal{N}_{f}(\cdot)$ to predict the shift $\Delta_{d}$ and focal length scaling factor $\alpha_{f}$ respectively. 
In our experiments we simply use an initial focal length with a field of view (FOV) of $60^{\circ}$. 
We have also tried to employ a single network to predict both the shift and the scaling factor, but have empirically found that two separate networks can achieve a better performance.

\subsection{Monocular Depth Prediction Module}
We train our depth prediction on multiple data sources including high-quality LiDAR sensor data~\cite{zamir2018taskonomy}, and low-quality web stereo data~\cite{Ranftl2020, wang2019web,  xian2020structure} (see Sec.~\ref{sec:data}).
As these datasets have varied depth ranges and web stereo datasets contain unknown depth scale and shift, we propose an image-level normalized regression (ILNR) loss to address this issue. 
Moreover, we propose a pair-wise normal regression (PWN) loss to improve local geometric features.

\paragraph{Image-level normalized regression loss.} 
Depth maps of different data sources can have varied depth ranges. Therefore, they need to be normalized to make the model training easier. Simple Min-Max normalization~\cite{ garcia2015data, singh2019investigating} is sensitive to depth value outliers. For example, a large value at a single pixel will affect the rest of the depth map after the Min-Max normalization. We investigate more robust normalization methods and propose a simple but effective image-level normalized regression loss for mixed-data training. 

Our image-level normalized regression loss transforms each ground truth depth map to a similar numerical range based on its individual statistics. 
To reduce the effect of outliers and long-tail residuals, we combine tanh normalization~\cite{singh2019investigating} with a trimmed Z-score, after which we can simply apply a pixel-wise mean average error (MAE) between the prediction and the normalized ground truth depth maps.
The ILNR loss is formally defined as follows.
\begin{align*}
\label{normalization loss}
  L_\text{ILNR} &= \frac{1}{N}{\sum_{i}^{N}\left|d_{i} - \overline{d}^{*}_{i}\right| + \left|\text{tanh}(\nicefrac{d_{i}}{100}) - \text{tanh}(\nicefrac{\overline{d}^{*}_{i}}{100}) \right|}
\end{align*}
where $\overline{d}^{*}_{i} =  \nicefrac{ ( d^{*}_{i} - \mu_\text{trim} )} {\sigma_\text{trim} }$ and 
$\mu_{\rm trim}$ and $\sigma_{\rm trim}$ are the mean and the standard deviation of a trimmed depth map which has the nearest and farthest $10\%$ of pixels removed, $d$ is the predicted depth, and $d^{*}$ is the ground truth depth map.
We have tested a number of other normalization methods such as Min-Max normalization~\cite{singh2019investigating}, Z-score normalization~\cite{fukunaga2013introduction}, and median absolute deviation normalization (MAD)~\cite{singh2019investigating}. In our experiments, we found that our proposed ILNR loss achieves the best performance.

\paragraph{Pair-wise normal loss.} 
Normals are an important geometric property, which have been shown to be a complementary modality to depth~\cite{silberman2012indoor}.
Many methods have been proposed to use normal constraints to improve the depth quality, such as the virtual normal loss~\cite{Yin2019enforcing}. 
However, as the virtual normal only leverages global structure, it cannot help improve the local geometric quality, such as depth edges and planes. 
Recently, Xian~\etal~\cite{xian2020structure} proposed a structure-guided ranking loss, which can improve edge sharpness.
Inspired by these methods, we follow their sampling method but enforce the supervision in surface normal space. 
Moreover, our samples include not only edges but also planes. 
Our proposed pair-wise normal (PWN) loss can better constrain both the global and local geometric relations. 

The surface normal is obtained from the reconstructed 3D point cloud by local least squares fitting~\cite{Yin2019enforcing}. Before calculating the predicted surface normal, we align the predicted depth and the ground truth depth with a scale and shift factor, which are retrieved by least squares fitting~\cite{Ranftl2020}. 
From the surface normal map, the planar regions where normals are almost the same and edges where normals change significantly can be easily located.
Then, we follow~\cite{xian2020structure} and sample paired points on both sides of these edges. If planar regions can be found, paired points will also be sampled on the same plane.
In doing so, we sample $~100$K paired points per training sample on average.
In addition, to improve the global geometric quality, we also randomly sample paired points globally.
The sampled points are $\{(A_{i}, B_{i}), i=0,...,N\}$, while their corresponding normals are $\{ (n_{Ai}, n_{Bi}), i=0,...,N \}$. 
The PWN loss is:
\begin{equation}
\label{normalization loss}
\begin{aligned}
  L_\text{PWN} &= \frac{1}{N}\sum_{i}^{N}\left|n_{Ai} \cdot n_{B_i} - n^{*}_{Ai}  \cdot  n^{*}_{Bi} \right|
\end{aligned}
\end{equation}
where $n^{*}$ denotes ground truth surface normals.
As this loss accounts for both local and global geometry, we find that it improves the overall reconstructed shape.

Finally, we also use a multi-scale gradient loss~\cite{li2018megadepth}:
\begin{equation}
L_\text{MSG}=\frac{1}{N}\sum_{k=1}^{K}\sum_{i=1}^{N}\left |\bigtriangledown ^{k}_{x}d_{i} -\bigtriangledown ^{k}_{x}\overline{d}^{*}_{i}    \right | + \left |\bigtriangledown ^{k}_{y}d_{i} -\bigtriangledown ^{k}_{y}\overline{d}^{*}_{i}    \right |
\label{multi-scale gradient loss}
\end{equation}
The overall loss function is formally defined as follows.
\begin{equation}     
L= L_\text{PWN} + \lambda_{a} L_\text{ILNR} + \lambda_{g} L_\text{MSG}
\label{overall loss}  
\end{equation}
where $\lambda_{a}=1$ and $\lambda_{g}=0.5$ in all experiments.

\section{Experiments}

\begin{table}[t]
\centering
\resizebox{1\linewidth}{!}{%
\begin{tabular}{l|llll}
\toprule[1pt]
\multirow{2}{*}{Dataset} & \multirow{2}{*}{\# Img} & Scene & Evaluation & Supervision \\
 &  & Type & Metric & Type \\ \hline
NYU & $654$ & Indoor & AbsRel  \& $\delta_{1}$ & Kinect \\ \hline
ScanNet & $700$ & Indoor & AbsRel \& $\delta_{1}$ & Kinect \\ \hline
2D-3D-S & $12256$ & Indoor & LSIV & LiDAR \\ \hline
\multirow{2}{*}{iBims-1} & \multirow{2}{*}{$100$} & \multirow{2}{*}{Indoor} & \multirow{2}{*}{\begin{tabular}[c]{@{}l@{}}AbsRel \&\\ $\varepsilon_\text{PE}$  \&$\varepsilon_\text{DBE}$\end{tabular}} & \multirow{2}{*}{LiDAR} \\
 &  &  &  &  \\ \hline 
KITTI & $652$ & Outdoor & AbsRel \& $\delta_{1}$ & LiDAR \\ \hline
Sintel & $641$ & Outdoor & AbsRel \& $\delta_{1}$ & Synthetic \\ \hline
ETH3D & $431$ & Outdoor & AbsRel \& $\delta_{1}$ & LiDAR \\ \hline 
YouTube3D & $58054$ & In the Wild & WHDR & SfM, Ordinal pairs \\  \hline
\multirow{2}{*}{OASIS} & \multirow{2}{*}{$10000$} & \multirow{2}{*}{In the Wild} & \multirow{2}{*}{\begin{tabular}[c]{@{}l@{}}WHDR \\ \& LSIV\end{tabular}} & \multirow{2}{*}{\begin{tabular}[c]{@{}l@{}}User clicks, \\ Small patches with GT\end{tabular}} \\
 &  &  &  &  \\ \hline
\multirow{2}{*}{DIODE} & \multirow{2}{*}{$771$} & \multirow{2}{*}{\begin{tabular}[c]{@{}l@{}}Indoor\\  \& Outdoor\end{tabular}} & \multirow{2}{*}{\begin{tabular}[c]{@{}l@{}}AbsRel \& \\ $\delta_{1}$\end{tabular}} & \multirow{2}{*}{LiDAR} \\
 &  &  &  &  \\ \toprule[1pt]
\end{tabular}}
\caption{Overview of the test sets in our experiments. \label{Tab: testing data details}}
\vspace{-0.5em}
\end{table}

\paragraph{Datasets and implementation details.}
\label{sec:data}
To train the PCM, we sampled $100$K Kinect-captured depth maps from ScanNet, $114$K LiDAR-captured depth maps from Taskonomy, and $51$K synthetic depth maps from the 3D Ken Burns paper~\cite{Niklaus_TOG_2019}.
We train the network using SGD with a batch size of $40$, an initial learning rate of $0.24$, and a learning rate decay of $0.1$. For parameters specific to PVCNN, such as the voxel size, we follow the original work~\cite{liu2019pvcnn}.

To train the DPM, we sampled $114$K RGBD pairs from LiDAR-captured Taskonomy~\cite{zamir2018taskonomy}, $51$K synthetic RGBD pairs from the 3D Ken Burns paper~\cite{Niklaus_TOG_2019},  $121$K RGBD pairs from calibrated stereo DIML~\cite{kim2018deep}, $48$K RGBD pairs from web-stereo Holopix50K~\cite{hua2020holopix50k}, and $20$K web-stereo HRWSI~\cite{xian2020structure} RGBD pairs.
Note that when doing the ablation study about the effectiveness of PWN and ILNR, we sampled a smaller dataset which is composed of $12$K images from Taskonomy, $12$K images from DIML, and $12$K images from HRWSI. 
During training, $1000$ images are withheld from all datasets as a validation set.
We use the depth prediction architecture proposed in Xian~\etal.~\cite{xian2020structure}, which consists of a standard backbone for feature extraction (e.g., ResNet50~\cite{he2016deep} or ResNeXt101~\cite{xie2017aggregated}), followed by a decoder, and train it using SGD with a batch size of $40$, an initial learning rate $0.02$ for all layer, and a learning rate decay of $0.1$. 
Images are resized to $448$ × $448$, and flipped horizontally with a $50\%$ chance.
Following \cite{yin2020diversedepth}, we load data from different datasets evenly for each batch.

\paragraph{Evaluation details.}
The focal length prediction accuracy is evaluated on 2D-3D-S~\cite{armeni2017joint} following~\cite{hold2018perceptual}. 
Furthermore, to evaluate the accuracy of the reconstructed 3D shape, we use the Locally Scale Invariant RMSE (LSIV)~\cite{chen2020oasis} metric on both OASIS~\cite{chen2020oasis} and 2D-3D-S~\cite{armeni2017joint}. It is consistent with the previous work~\cite{chen2020oasis}. The OASIS~\cite{chen2020oasis} dataset only has the ground truth depth on some small regions, while 2D-3D-S has the ground truth for the whole scene.

\begin{table}[t]
\centering
\resizebox{1\linewidth}{!}{%
\begin{tabular}{l|ccccc}
\toprule[1pt]
\multicolumn{1}{c|}{\multirow{2}{*}{Method}} & ETH3D & NYU & KITTI & Sintel & DIODE \\
\multicolumn{1}{c|}{}                        & \multicolumn{5}{c}{AbsRel $\downarrow$}           \\ \hline
Baseline    & $23.7$    &$25.8$     &$23.3$   &$47.4$   &$46.8$    \\
Recovered Shift  & $\textbf{15.9}$    &$\textbf{15.1}$    &$\textbf{17.5}$  &$\textbf{40.3}$   &$\textbf{36.9}$  \\ \toprule[1pt]
\end{tabular}}
\caption{Effectiveness of recovering the shift from 3D point clouds with the PCM. Compared with the baseline, the AbsRel is much lower after recovering the depth shift over all test sets.\label{Tab: effectiveness of shift prediction. }}
\vspace{-0.5em}
\end{table}

\begin{figure}[t]   
\centering
\includegraphics[width=\linewidth]{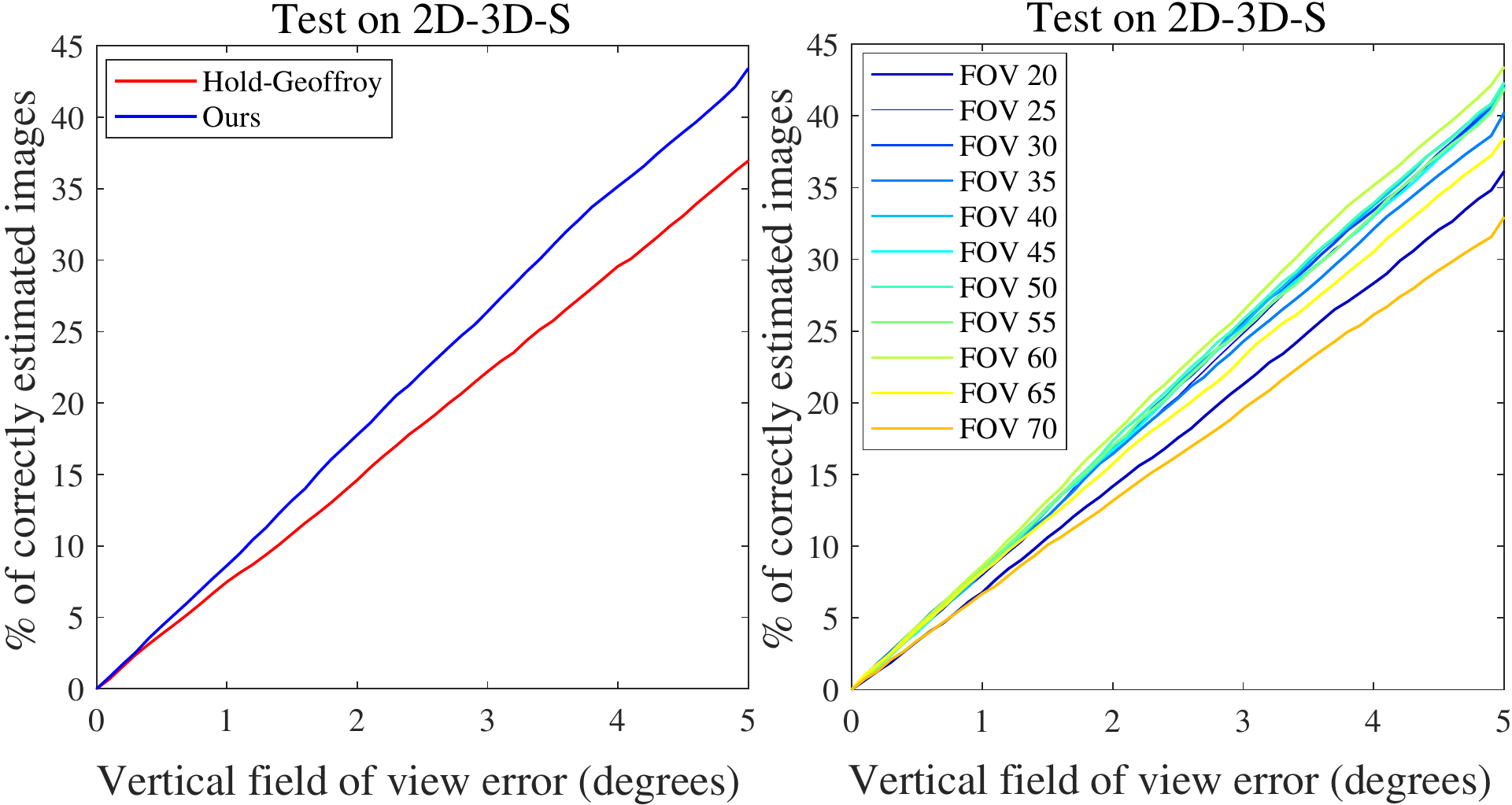}
\caption{Comparison of recovered focal length on the 2D-3D-S dataset. Left, our method outperforms Hold-Geoffroy~\etal~\cite{hold2018perceptual}. Right, we conduct an experiment on the effect of the initialization of field of view (FOV). Our method remains robust across different initial FOVs, with a slight degradation in quality past $25^{\circ}$ and $65^{\circ}$.}
\vspace{-0.5em}
\label{Fig: cmp of focal length prediction.}
\end{figure}

\begin{figure*}[t]
\centering
\includegraphics[width=\linewidth]{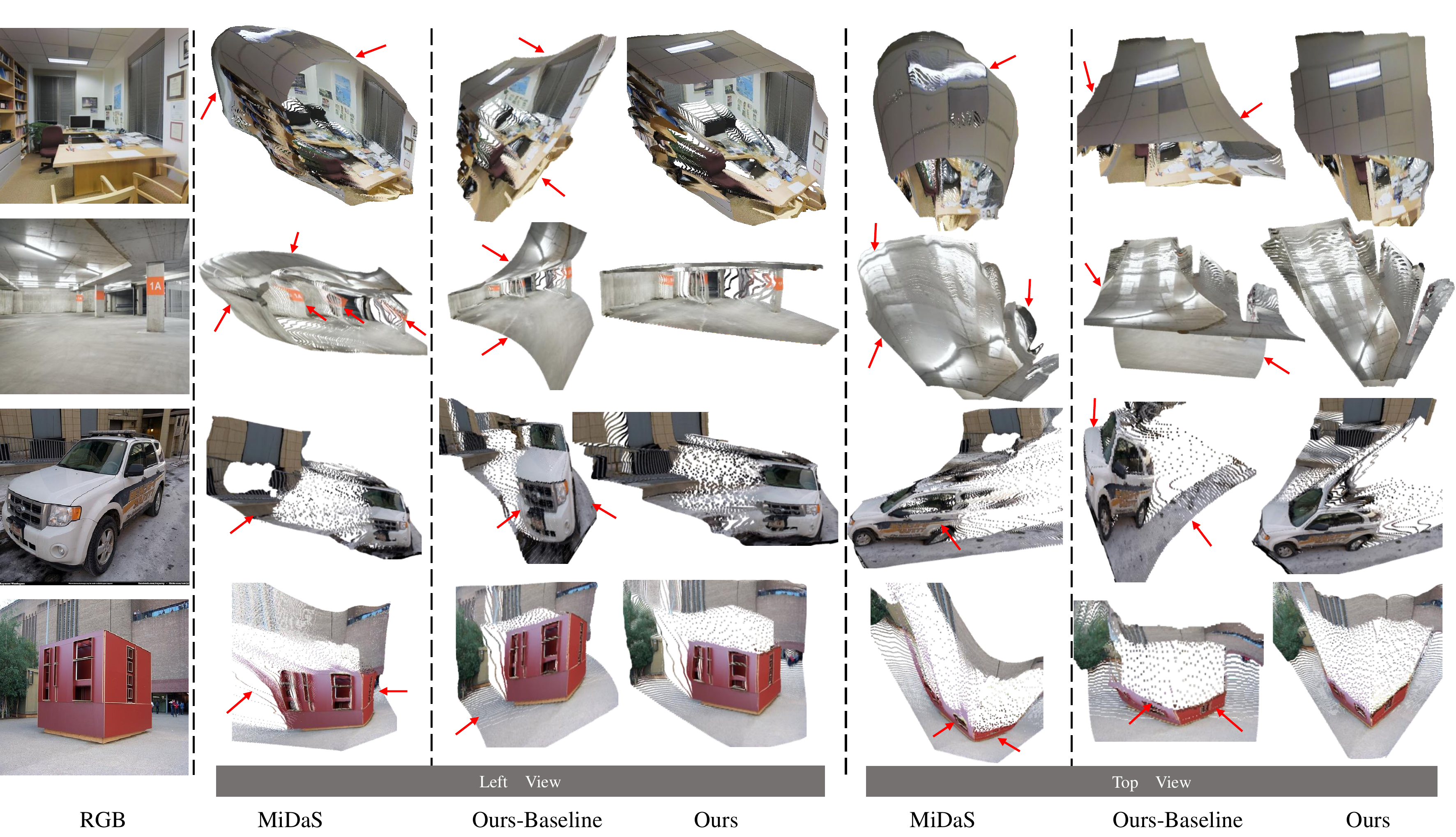}
\caption{Qualitative comparison. We compare the reconstructed 3D shape of our method with several baselines. As MiDaS~\cite{Ranftl2020} does not estimate the focal length, we use the focal length recovered from \cite{hold2018perceptual} to convert the predicted depth to a point cloud.
``Ours-Baseline'' does not recover the depth shift or focal length and uses an orthographic camera, while ``Ours'' recovers the shift and focal length. We can see that our method better reconstructs the 3D shape, especially at edges and planar regions (see arrows).}
\label{Fig: cmp of SOTA 3D shape. }
\vspace{-0.5em}
\end{figure*}

To evaluate the generalizability of our proposed depth prediction method, we take $9$ datasets which are unseen during training, including YouTube3D~\cite{chen2019learning}, OASIS~\cite{chen2020oasis}, NYU~\cite{silberman2012indoor}, KITTI~\cite{geiger2012we}, ScanNet~\cite{dai2017scannet}, DIODE~\cite{vasiljevic2019diode}, ETH3D~\cite{schops2017multi}, Sintel~\cite{Butler:ECCV:2012}, and iBims-1~\cite{Koch18:ECS}. 
On OASIS and YouTube3D, we use the Weighted Human Disagreement Rate (WHDR)~\cite{xian2018monocular} for evaluation. 
On other datasets, except for iBims-1, we evaluate the absolute mean relative error (AbsRel) and the percentage of pixels with $\delta_{1}=\text{max}(\frac{d_{i}}{d^{*}_{i}}, \frac{d^{*}_{i}}{d_{i}})<1.25$. 
We follow Ranftl~\etal~\cite{Ranftl2020} and align the scale and shift before evaluation. 
To evaluate the geometric quality of the depth, i.e. the quality of edges and planes, we follow~\cite{Niklaus_TOG_2019, xian2020structure} and evaluate the depth boundary error~\cite{Koch18:ECS} ($\varepsilon^\text{acc}_\text{DBE}, \varepsilon^\text{comp}_\text{DBE}$) as well as the planarity error~\cite{Koch18:ECS} ($\varepsilon^\text{plan}_\text{PE}, \varepsilon^\text{orie}_\text{PE}$) on iBims-1. 
$\varepsilon^\text{plan}_\text{PE}$ and $\varepsilon^\text{orie}_\text{PE}$ evaluate the flatness and orientation
of reconstructed 3D planes compared to the ground truth 3D planes respectively, while $\varepsilon^\text{acc}_\text{DBE}$ and $\varepsilon^\text{comp}_\text{DBE}$ demonstrate the localization accuracy and the sharpness of edges respectively. More details as well as a comparison of these test datasets are summarized in Tab.~\ref{Tab: testing data details}

\subsection{3D Shape Reconstruction}

\paragraph{Shift recovery.}
To evaluate the effectiveness of our depth shift recovery, we perform zero-shot evaluation on $5$ datasets unseen during training.
We recover a 3D point cloud by unprojecting the predicted depth map, and then compute the depth shift using our PCM. 
We then align the unknown scale~\cite{bian2019unsupervised, monodepth2} of the original depth and our shifted depth to the ground truth, and evaluate both using the AbsRel error.
The results are shown in Tab.~\ref{Tab: effectiveness of shift prediction. }, where we see that, on all test sets, the AbsRel error is lower after recovering the shift. 
We also trained a standard 2D convolutional neural network to predict the shift given an image composed of the unprojected point coordinates, but this approach did not generalize well to samples from unseen datasets. 

\paragraph{Focal length recovery.}

To evaluate the accuracy of our recovered focal length, we follow Hold-Geoffroy~\etal~\cite{hold2018perceptual} and compare on the 2D-3D-S dataset, which is unseen during training for both methods. 
The model of~\cite{hold2018perceptual} is trained on the in-the-wild SUN360~\cite{xiao2012recognizing} dataset. 
Results are illustrated in Fig.~\ref{Fig: cmp of focal length prediction.}, where we can see that our method demonstrates better generalization performance.
Note that PVCNN is very lightweight and only has $5.5M$ parameters, but shows promising generalizability, which could indicate that there is a smaller domain gap between datasets in the 3D point cloud space than in the image space where appearance variation can be large.

Furthermore, we analyze the effect of different initial focal lengths during inference. 
We set the initial field of view (FOV) from $20^{\circ}$ to $70^{\circ}$ and evaluate the accuracy of the recovered focal length, Fig.~\ref{Fig: cmp of focal length prediction.} (right). 
The experimental results demonstrate that our method is not particularly sensitive to different initial focal lengths.

\begin{table}[t]
\centering
\resizebox{\linewidth}{!}{%
\begin{tabular}{l|ll}
\toprule
\multirow{2}{*}{Method} & OASIS & 2D-3D-S \\
                        & LSIV $\downarrow$ & LSIV$\downarrow$  \\ \hline \hline  
\multicolumn{3}{c}{Orthographic Camera Model} \\ \hline \hline               
MegaDepth~\cite{li2018megadepth}               &$0.64$       &$2.68$       \\
MiDaS~\cite{Ranftl2020}               &$0.63$       &$2.65$       \\
Ours-DPM              &$0.63$       &$2.65$      \\ \hline \hline
\multicolumn{3}{c}{Pinhole Camera Model} \\ \hline \hline 
MegaDepth~\cite{li2018megadepth} + Hold-Geoffroy~\cite{hold2018perceptual} &$1.69$  &$1.81$      \\
MiDaS~\cite{Ranftl2020} + Hold-Geoffroy~\cite{hold2018perceptual}&$1.60$ &$0.94$       \\
MiDaS~\cite{Ranftl2020} + Ours-PCM &$1.32$ &$0.94$       \\
Ours                   &$\textbf{0.52}$       &$\textbf{0.80}$       \\ 
\bottomrule
\end{tabular}}
\vspace{0.4em}
\caption{Quantitative evaluation of the reconstructed 3D shape quality on OASIS and 2D-3D-S. Our method can achieve better performance than previous methods. Compared with the orthographic projection, our method using the pinhole camera model can obtain better performance. DPM and PCM refers to our depth prediction module and point cloud module respectively.
\label{Tab: shape evaluation on OASIS}}
\vspace{-2em}
\end{table}

\begin{figure*}[t]
\centering
\includegraphics[width=\linewidth]{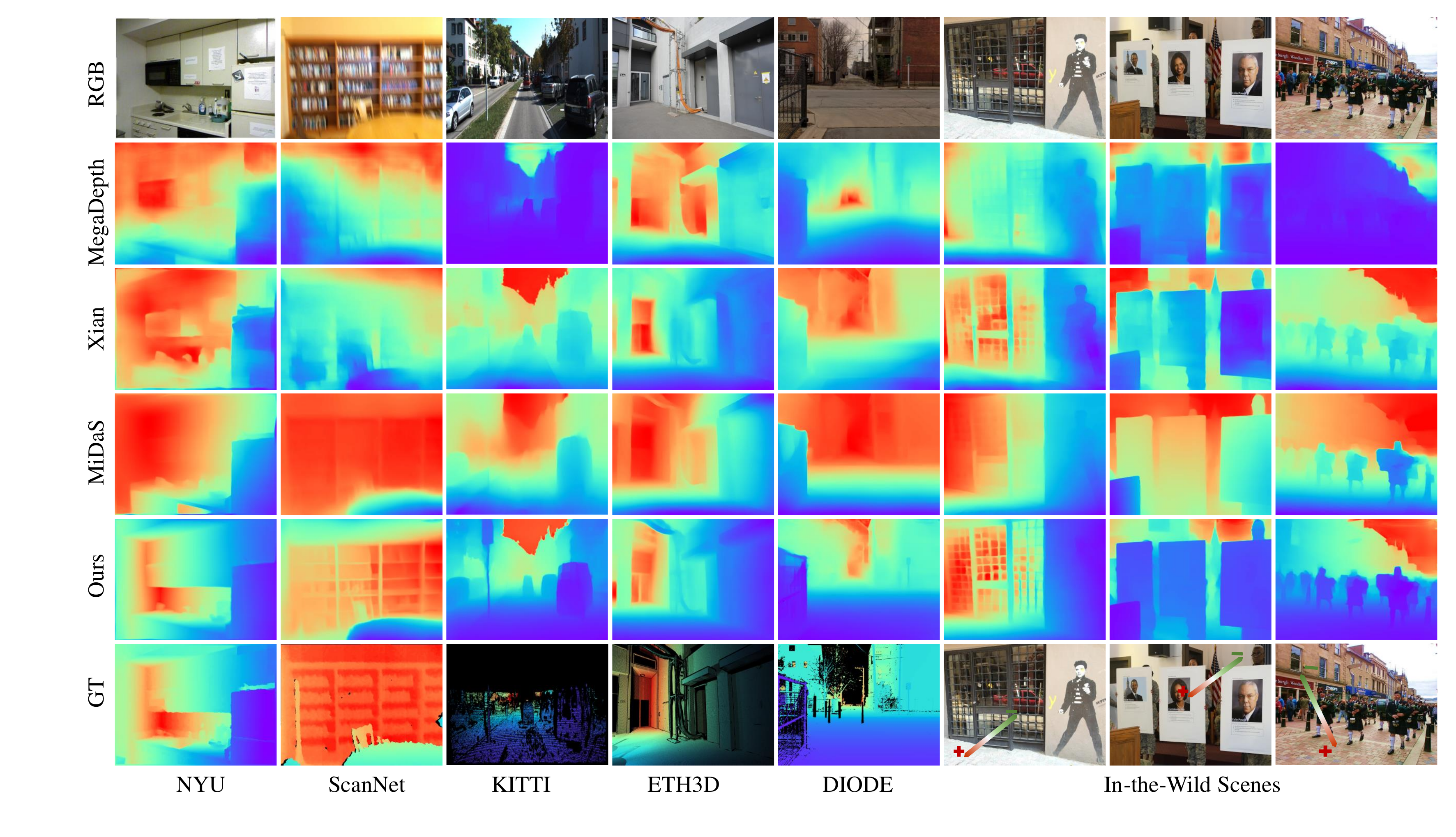}
\caption{Qualitative comparisons with state-of-the-art methods, including MegaDepth~\cite{li2018megadepth}, Xian~\etal\cite{xian2020structure}, and MiDaS~\cite{Ranftl2020}. It shows that our method can predict more accurate depths at far locations and regions with complex details. In addition, we see that our method generalizes better on in-the-wild scenes.\label{Fig: cmp of SOTA. }}
\vspace{-0.5em}
\end{figure*}

\paragraph{Evaluation of 3D shape quality.}
Following OASIS~\cite{chen2020oasis}, we use LSIV for the quantitative comparison of recovered 3D shapes on the OASIS~\cite{chen2020oasis} dataset and the 2D-3D-S~\cite{armeni2017joint} dataset. 
OASIS only provides the ground truth point cloud on small regions, while 2D-3D-S covers the whole 3D scene. %
Following OASIS~\cite{chen2020oasis}, we evaluate the reconstructed 3D shape with two different camera models, i.e. the orthographic projection camera model~\cite{chen2020oasis} (infinite focal length) and the (more realistic) pinhole camera model. 
As MiDaS~\cite{Ranftl2020} and MegaDepth~\cite{li2018megadepth} do not estimate the focal length, we use the focal length recovered from Hold-Geoffroy~\cite{hold2018perceptual} to convert the predicted depth to a point cloud.
We also evaluate a baseline using MiDaS instead of our DPM with the focal length predicted by our PCM (``MiDaS + Ours-PCM'').
From Tab.~\ref{Tab: shape evaluation on OASIS} we can see that with an orthographic projection, our method (``Ours-DPM'') performs  roughly as well as existing state-of-the-art methods. However, for the pinhole camera model our combined method significantly outperforms existing approaches. Furthermore, comparing ``MiDaS + Ours-PCM'' and ``MiDaS + Hold-Geoffroy'', we note that our PCM is able to generalize to different depth prediction methods. 

A qualitative comparison of the reconstructed 3D shape on in-the-wild scenes is shown in Fig.~\ref{Fig: cmp of SOTA 3D shape. }. 
It demonstrates that our model can recover more accurate 3D scene shapes. For example, planar structures such as walls, floors, and roads are much flatter in our reconstructed scenes, and the angles between surfaces (\eg walls) are also more realistic. Also, 
the shape of the car has less distortions.
\begin{table}[t]
\small
\centering
\begin{threeparttable}
\resizebox{1\linewidth}{!}{%
\begin{tabular}{l|ccccc}
\toprule[1pt]
\multirow{2}{*}{Method} & \multicolumn{4}{c}{iBims-1}                                    \\ \cline{2-6} 
        & $\varepsilon^\text{acc}_\text{DBE}\downarrow$ & \multicolumn{1}{l|}{$\varepsilon^\text{comp}_\text{DBE}\downarrow$} & $\varepsilon^\text{plan}_\text{PE}\downarrow$ & \multicolumn{1}{l|}{$\varepsilon^\text{orie}_\text{PE}\downarrow$} &AbsRel$\downarrow$\\ \hline
Xian~\cite{xian2020structure}         &$7.72$  & \multicolumn{1}{l|}{$9.68$}         &$5.00$ & \multicolumn{1}{l|}{$44.77$}   &$0.301$    \\
MegaDepth~\cite{li2018megadepth}    & $4.09$         & \multicolumn{1}{l|}{$8.28$} & $7.04$ &\multicolumn{1}{l|}{ $33.03$} &$0.20$ \\
MiDaS~\cite{Ranftl2020}        &$1.91$ & \multicolumn{1}{l|}{$5.72$} &$3.43$ & \multicolumn{1}{l|}{$12.78$}    &$0.104$    \\

3D Ken Burns~\cite{Niklaus_TOG_2019}   & $2.02$ & \multicolumn{1}{l|}{$\textbf{5.44}$} & $\underline{2.19}$   & \multicolumn{1}{l|}{$\underline{10.24}$} & $\underline{0.097}$  \\
\hline \hline
Ours\tnote{\dag} \, w/o PWN  & $2.05$ & \multicolumn{1}{l|}{$6.10$}  & $3.91$  & \multicolumn{1}{l|}{$13.47$} & $0.106$  \\
Ours\tnote{\dag}    & $\underline{1.91}$ & \multicolumn{1}{l|}{$\underline{5.70}$}  & $2.95$  & \multicolumn{1}{l|}{$11.59$} & $0.101$       \\
Ours Full   & $\textbf{1.90}$ & \multicolumn{1}{l|}{$5.73$}  & $\textbf{2.0}$  & \multicolumn{1}{l|}{$\textbf{7.41}$}  & $\textbf{0.079}$      \\
\bottomrule
\end{tabular}}
\end{threeparttable}
\vspace{0.4em}
\caption{Quantitative comparison of the quality of depth boundaries (DBE) and planes (PE) on the iBims-1 dataset. We use $^\dag$ to indicate when a method was trained on the small training subset.\label{Tab: cmp of edges and planes}}
\vspace{-2em}
\end{table}

\subsection{Depth prediction}
In this section, we conduct several experiments to demonstrate the effectiveness of our depth prediction method, including a comparison with state-of-the-art methods, a comparison of our proposed image-level normalized regression loss with other methods, and an analysis of the effectiveness of our pair-wise normal regression loss. 

\begin{table*}[t]
\setlength{\tabcolsep}{2pt}
\resizebox{\linewidth}{!}{%
\begin{tabular}{l|l|ll|ll|ll|ll|ll|ll|ll|l}
\toprule[1pt]
\multirow{2}{*}{Method} & \multirow{2}{*}{Backbone} & OASIS & YT3D & \multicolumn{2}{c|}{NYU} & \multicolumn{2}{c|}{KITTI} & \multicolumn{2}{c|}{DIODE} & \multicolumn{2}{c|}{ScanNet} & \multicolumn{2}{c|}{ETH3D} & \multicolumn{2}{c|}{Sintel} & \multirow{2}{*}{Rank} \\
&   &\multicolumn{2}{c|}{WHDR$\downarrow$}    & AbsRel$\downarrow$     & $\delta_{1}\uparrow$     & AbsRel$\downarrow$      & $\delta_{1}\uparrow$      & AbsRel$\downarrow$      & $\delta_{1}\uparrow$      &AbsRel$\downarrow$      & $\delta_{1}\uparrow$       &AbsRel$\downarrow$     & $\delta_{1}\uparrow$      & AbsRel$\downarrow$       & $\delta_{1}\uparrow$      &                       \\ \hline
OASIS~\cite{chen2020oasis}  &ResNet50  & $32.7$ &$27.0$ &$21.9$ &$66.8$ &$31.7$ & $43.7$ &  $48.4$ &$53.4$ &$19.8$ &$69.7$ &$29.2$ &$59.5$ &$60.2$ &$42.9$  &  $6.7$\\ 
MegaDepth~\cite{li2018megadepth}& Hourglass &$33.5$  &$26.7$  &$19.4$& $71.4$ &$20.1$ &$66.3$ &$39.1$ &$61.5$ &$19.0$ &$71
.2$ &$26.0$  &$64.3$ &$39.8$ &$52.7$  &$6.7$ \\
Xian~\cite{xian2020structure} &ResNet50 &$31.6$ &$23.0$ &$16.6$ &$77.2$ & $27.0$  & $52.9$ &$42.5$ &$61.8$ &$17.4$ &$75.9$ &$27.3$ &$63.0$ &$52.6$ &$50.9$ & $6.7$    \\
WSVD~\cite{wang2019web} &ResNet50 &$34.8$  &$24.8$  &$22.6$ &$65.0$ &$24.4$ &$60.2$ &$35.8$ &$63.8$ &$18.9$ &$71.4$ &$26.1$ &$61.9$ &$35.9$  & $54.5$ &$6.6$ \\
Chen~\cite{chen2019learning} &ResNet50  &$33.6$ &$20.9$  &  $16.6$ & $77.3$ & $32.7$ & $51.2$ &$37.9$ & $66.0$ & $16.5$ & $76.7$  &$23.7$ &$67.2$ &$38.4$ & $57.4$ &$5.6$    \\
DiverseDepth~\cite{yin2020diversedepth}&ResNeXt50 &$30.9$ &$21.2$ &$11.7$ &$87.5$ &$19.0$ &$70.4$ &$37.6$ &$63.1$ &$10.8$ &$88.2$ &$22.8$ &$69.4$ &$38.6$ &$58.7$  &$4.4$  \\
MiDaS~\cite{Ranftl2020}&ResNeXt101 &$\underline{29.5}$ &$19.9$ &$11.1$ &$88.5$ &$23.6$ &$63.0$ &$33.2$ &$71.5$ &$11.1$ &$88.6$  & $18.4$ &$75.2$ &$40.5$ &$60.6$ & $3.5$ \\
\hline
Ours &ResNet50 &$30.2$ &$\underline{19.5}$  &$\underline{9.1}$  &$\underline{91.4}$  &$\textbf{14.3}$ &$\textbf{80.0}$ &$\underline{28.7}$ &$\underline{75.1}$ &$\underline{9.6}$ &$\underline{90.8}$ &$\underline{18.4}$ &$\underline{75.8}$ &$\underline{34.4}$ &$\underline{62.4}$ &$\underline{1.9}$   \\     
Ours &ResNeXt101 &$\textbf{28.3}$ &$\textbf{19.2}$  &$\textbf{9.0}$  &$\textbf{91.6}$  &$\underline{14.9}$ &$\underline{78.4}$ &$\textbf{27.1}$ &$\textbf{76.6}$ &$\textbf{9.5}$ &$\textbf{91.2}$ &$\textbf{17.1}$ &$\textbf{77.7}$ &$\textbf{31.9}$ &$\textbf{65.9}$ &$\textbf{1.1}$  
\\ \toprule[1pt]
\end{tabular}}
\caption{Quantitative comparison of our depth prediction with state-of-the-art methods on eight zero-shot (unseen during training) datasets. Our method achieves better performance than existing state-of-the-art methods across all test datasets. \label{Tab: Cmp with SOTA}}
\vspace{-1.5em}
\end{table*}

\paragraph{Comparison with state-of-the-art methods.}
In this comparison, we test on datasets unseen during training.
We compare with methods that have been shown to best generalize to in-the-wild scenes. Their results are obtained by running the publicly released code. Each method is trained on its own proposed datasets.
When comparing the AbsRel error, we follow Ranftl~\cite{Ranftl2020} to align the scale and shift before the evaluation. 
The results are shown in the Tab.~\ref{Tab: Cmp with SOTA}. Our method outperforms prior works, and using a larger ResNeXt101 backbone further improves the results. Some qualitative comparisons can be found in Fig.~\ref{Fig: cmp of SOTA. }

\begin{figure}[t]
\centering
\includegraphics[width=\linewidth]{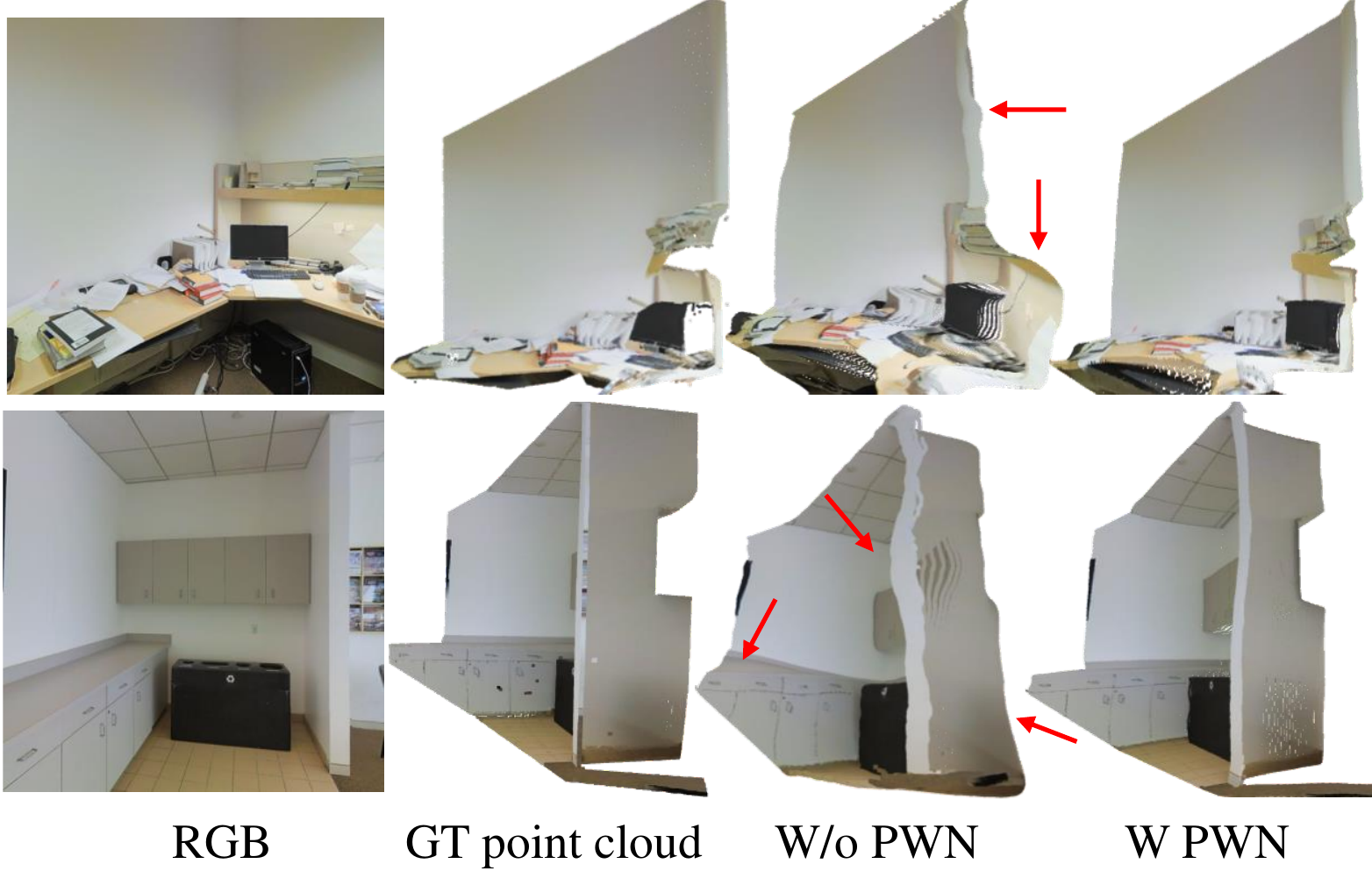}
\caption{Qualitative comparison of reconstructed point clouds. Using the pair-wise normal loss (PWN), we can see that edges and planes are better reconstructed (see highlighted regions).}
\label{Fig: cmp of pcd with PWN. }
\vspace{-1em}
\end{figure}

 \paragraph{Pair-wise normal loss.} 
 To evaluate the quality of our full method and dataset on edges and planes, we compare our depth model with previous state-of-the-art methods on the iBims-1 dataset.
 In addition, we evaluate the effect of our proposed pair-wise normal (PWN) loss through an ablation study. As training on our full dataset is computationally demanding, we perform this ablation on the small training subset.
 The results are shown in Tab.~\ref{Tab: cmp of edges and planes}. 
 We can see that our full method outperforms prior work for this task.
 In addition, under the same settings, both edges and planes are improved by adding the PWN loss. 
 We further show a qualitative comparison in Fig.~\ref{Fig: cmp of pcd with PWN. }.

\paragraph{Image-level normalized regression loss.}
To show the effectiveness of our proposed image-level normalized regression (ILNR) loss, we compare it with the scale-shift invariant loss (SSMAE)~\cite{Ranftl2020} and the scale-invariant multi-scale gradient loss~\cite{wang2019web}. 
Each of these methods is trained on the small training subset to limit the computational overhead, and comparisons are made to datasets that are unseen during training. 
All models have been trained for $50$ epochs, and we have verified that all models fully converged by then. 
The quantitative comparison is shown in Tab.~\ref{Tab. cmp of normalization loss}, where we can see an improvement of ILNR over other scale and shift invariant losses. 
Furthermore, we also analyze different options for normalization, including image-level Min-Max (ILNR-MinMax) normalization and image-level median absolute deviation (ILNR-MAD) normalization, and found that our proposed loss performs a bit better.

\section{Discussion}

\paragraph{Limitations.}
We observed a few limitations of our method. 
For example, our PCM cannot recover accurate focal length or depth shift when the scene does not have enough geometric cues, \eg when the whole image is mostly a wall or a sky region. 
The accuracy of our method will also decrease with images taken from uncommon view angles (e.g., top-down) or extreme focal lengths. 
More diverse 3D training data may address these failure cases. 
In addition, our method does not model the effect of radial distortion from the camera and thus the reconstructed scene shape can be distorted in cases with severe radial distortion. 
Studying how to recover the radial distortion parameters using our PCM can be an interesting future direction.

\begin{table}[t]
\resizebox{\linewidth}{!}{%
\begin{tabular}{l|c|cccc}
\toprule[1pt]
Method     & RedWeb & NYU & KITTI & ScanNet & DIODE \\
           & WHDR$\downarrow$   & \multicolumn{4}{c}{AbsRel$\downarrow$}    \\ \hline
SMSG~\cite{wang2019web} & $19.1$ & $15.6$ &$16.3$ &$13.7$ & $36.5$\\
SSMAE~\cite{Ranftl2020} & $19.2$ & $14.4$ &$18.2$ &$13.3$ & $34.4$\\\hline  \hline
ILNR-MinMax  & $19.1$ &$15.0$ & $17.1$ & $13.3$ & $46.1$ \\
ILNR-MAD   &$18.8$  &$14.8$ &$17.4$ &$12.5$  & $34.6$  \\ \hline  \hline
ILNR  &$\textbf{18.7}$ & $\textbf{13.9}$ & $\textbf{16.1}$ & $\textbf{12.3}$ &$\textbf{34.2}$ \\ \toprule[1pt]
\end{tabular}}
\caption{Quantitative comparison of different losses on zero shot generalization to $5$ datasets unseen during training.\label{Tab. cmp of normalization loss}}
\vspace{-2em}
\end{table}

\section{Conclusion}

In summary, we presented, to our knowledge, the first fully data driven method that reconstructs 3D scene shape from a monocular image. 
To recover the shift and focal length for 3D reconstruction, we proposed to use point cloud networks trained on datasets with known global depth shifts and focal lengths.
This approach showed strong generalization capabilities and we are under the impression that it may be helpful for related depth-based tasks. 
Extensive experiments demonstrated the effectiveness of our scene shape reconstruction method and the superior ability to generalize to unseen data.
%

%
%

%
%
\let\mathscr=\mathcal

\appendix

\section*{Appendix}

\section{Datasets}
\subsection{Datasets for Training}
To train a robust model, we use a variety of data sources, each with its own unique properties:
\begin{itemize}
    \item Taskonomy~\cite{zamir2018taskonomy} contains high-quality RGBD data captured by a LiDAR scanner. We sampled around $114$K RGBD pairs for training.
    \item DIML~\cite{kim2018deep} contains calibrated stereo images. We use the GA-Net~\cite{Zhang2019GANet} method to compute the disparity for supervision. We sampled around $121$K RGBD pairs for training.
    \item 3D Ken Burns~\cite{Niklaus_TOG_2019} contains synthetic data with ground truth depth. We sampled around $51$K RGBD pairs for training.
    \item Holopix50K~\cite{hua2020holopix50k} contains diverse uncalibrated web stereo images. Following~\cite{xian2018monocular}, we use FlowNet~\cite{IMKDB17} to compute the relative depth (inverse depth) data for training.
    \item HRWSI~\cite{xian2020structure} contains diverse uncalibrated web stereo images. We use the entire dataset, consisting of $20$K RGBD images.
\end{itemize}

\subsection{Datasets Used in Testing}
To evaluate the generalizability of our method, we test our depth model on a range of datasets:
\begin{itemize}
    \item NYU~\cite{silberman2012indoor} consists of mostly indoor RGBD images where the depth is captured by a Kinect sensor. We test our method on the official test set, which contains $654$ images.
    \item KITTI~\cite{geiger2012we} consists of street scenes, with sparse metric depth captured by a LiDAR sensor. We use the standard test set ($652$ images) of the Eigen split.
    \item ScanNet~\cite{dai2017scannet} contains similar data to NYU, indoor scenes captured by a Kinect. We randomly sampled $700$ images from the official validation set for testing.
    \item DIODE~\cite{vasiljevic2019diode} contains high-quality LiDAR-generated depth maps of both indoor and outdoor scenes. We use the whole validation set ($771$ images) for testing.
    \item ETH3D~\cite{schops2017multi} consists of outdoor scenes whose depth is captured by a LiDAR sensor. We sampled $431$ images from it for testing.
    \item Sintel~\cite{Butler:ECCV:2012} is a synthetic dataset, mostly with outdoor scenes. We collected $641$ images from it for testing.
    \item OASIS~\cite{chen2020oasis} is a diverse dataset consisting of images in the wild, with ground truth depth annotations by humans. It contains both sparse relative depth labels (similar to DIW~\cite{chen2016single}), and some planar regions. We test on the entire validation set, containing $10$K images. 
    \item YouTube3D~\cite{chen2019learning} consists of in-the-wild videos that are reconstructed using structure from motion, with the sparse reconstructed points as supervision. We randomly sampled $58$K images from the whole dataset for testing.
    \item RedWeb~\cite{xian2018monocular} consists of in-the-wild stereo images, with disparity labels derived from an optical flow matching algorithm. We use $3.6$K images to evaluate the WHDR error, and we randomly sampled $5$K points pairs on each image.
    \item iBims-1~\cite{Koch18:ECS} is an indoor-scene dataset, which consists of $100$ high-quality images captured by a LiDAR sensor. We use the whole dataset for evaluating edge and plane quality.
\end{itemize}
We will release a list of all images used for testing to facilitate reproducibility.

\section{Details for Depth Prediction Model and Training.}
We use the depth prediction model proposed by Xian~\etal~\cite{xian2020structure}.
We follow~\cite{yin2020diversedepth} and combine the multi-source training data by evenly sampling from all sources per batch.
As HRWSI and Holopix50K are both web stereo data, we merge them together. Therefore, there are four different data sources, i.e. high-quality Taskonomy, synthetic 3D Ken Burn, middle-quality DIML, and low-quality Holopix50K and HRWSI. For example, if the batch size is $8$, we sample $2$ images from each of the four sources.
Furthermore, as the ground truth depth quality varies between data sources, we enforce different losses for them.

For the web-stereo data, such as Holopix50K~\cite{hua2020holopix50k} and HRWSI~\cite{xian2020structure}, as their inverse depths have unknown scale and shift, these inverse depths cannot be used to compute the affine-invariant depth (up to an unknown scale and shift to the metric depth).
The pixel-wise regression loss and geometry loss cannot be applied for such data. Therefore, during training, we only enforce the ranking loss~\cite{xian2018monocular} on them.

For the middle-quality calibrated stereo data, such as DIML~\cite{kim2018deep}, we enforce the proposed image-level normalized regression loss, multi-scale gradient loss and ranking loss. As the recovered disparities contain much noise in local regions, enforcing the pair-wise normal regression loss on noisy edges will cause many artifacts. Therefore, we enforce the pair-wise normal regression loss only on planar regions for this data.

For the high-quality data, such as Taskonomy~\cite{zamir2018taskonomy} and synthetic 3D Ken Burns~\cite{Niklaus_TOG_2019}, accurate edges and planes can be located. Therefore, we apply the pair-wise normal regression loss, ranking loss, and multi-scale gradient loss for this data.

\begin{figure}[t]
\centering
\includegraphics[width=\linewidth]{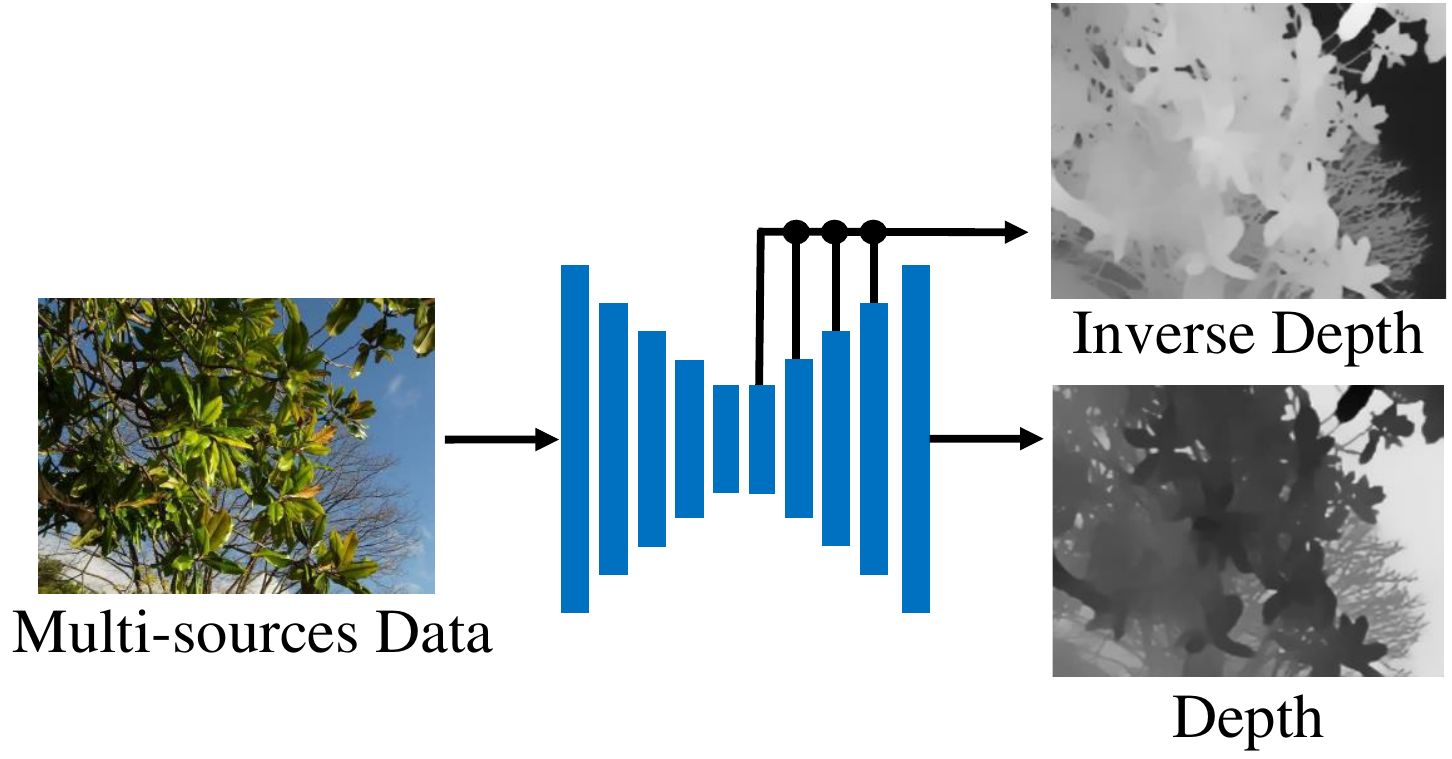}
\caption{The network architecture for the DPM. The network has two output branches. The decoder outputs the depth map, while the auxiliary path outputs the inverse depth. Different losses are enforced on these two branches.}
\label{Fig: network. }
\vspace{-0.5em}
\end{figure}

Furthermore, we follow \cite{liu2019training} and add a light-weight auxiliary path on the decoder.
The auxiliary outputs the inverse depth and the main branch (decoder) outputs the depth.
For the auxiliary path, we enforce the ranking loss, image-level normalized regression loss in the inverse depth space on all data.
The network is illustrated in Fig.~\ref{Fig: network. }.

\section{Sampling Strategy for Pairwise Normal Loss}
We enforce the pairwise normal regression loss on Taskonomy and DIML data. As DIML is more noisy than Taskonomy, we only enforce the normal regression loss on the planar regions, such as pavements and roads, whereas for Taskonomy, we sample points on edges and on planar regions.
We use the local least squares fitting method~\cite{Yin2019enforcing} to compute the surface normal from the depth map.

For edges, we follow the method of Xian~\etal~\cite{xian2020structure}, which we describe here.
The first step is to locate image edges. At each edge point, we then sample pairs of points on both sides of the edge, i.e. $\mathscr{P} = \{(P_{A}, P_{B})_{i} | i=0,...,n\}$.
The ground truth normals for these points are $\mathscr{N}^{*} = \{(\bm{n}^{*}_{A}, \bm{n}^{*}_{B})_{i} | i=0,...,n\}$, while the predicted normals are $\mathscr{N} = \{(\bm{n}_{A}, \bm{n}_{B})_{i} | i=0,...,n\}$.
To locate the object boundaries and planes folders, where the normals changes significantly, we set the angle difference of two normals greater than $arccos(0.3)$. To balance the samples, we also get some negative samples, where the angle difference is smaller than  $arccos(0.95)$ and they are also detected as edges. The sampling method is illustrated as follow.
\begin{equation}
    \mathscr{S}_{1} = \{\bm{n}^{*}_{A} \cdot \bm{n}^{*}_{B} > 0.95, \bm{n}^{*}_{A} \cdot \bm{n}^{*}_{B} < 0.3  | {(\bm{n}^{*}_{A}, \bm{n}^{*}_{B})_{i}} \in \mathscr{N}^{*} \}
\label{eq: 3D point cloud reconstruction}
\end{equation}

For planes, on DIML, we use \cite{deeplabv3plus2018} to segment the roads, which we assume to be planar regions. On Taskonmy, we locate planes by finding regions with the same normal.
On each detected plane, we sample $5000$ paired points.
Finally, we combine both sets of  paired points and enforce the normal regression loss on them, see E.q. $4$ in our main paper.

\section{Illustration of the Reconstructed Point Cloud}
\begin{figure*}
{
	\includegraphics[width=0.9\linewidth]{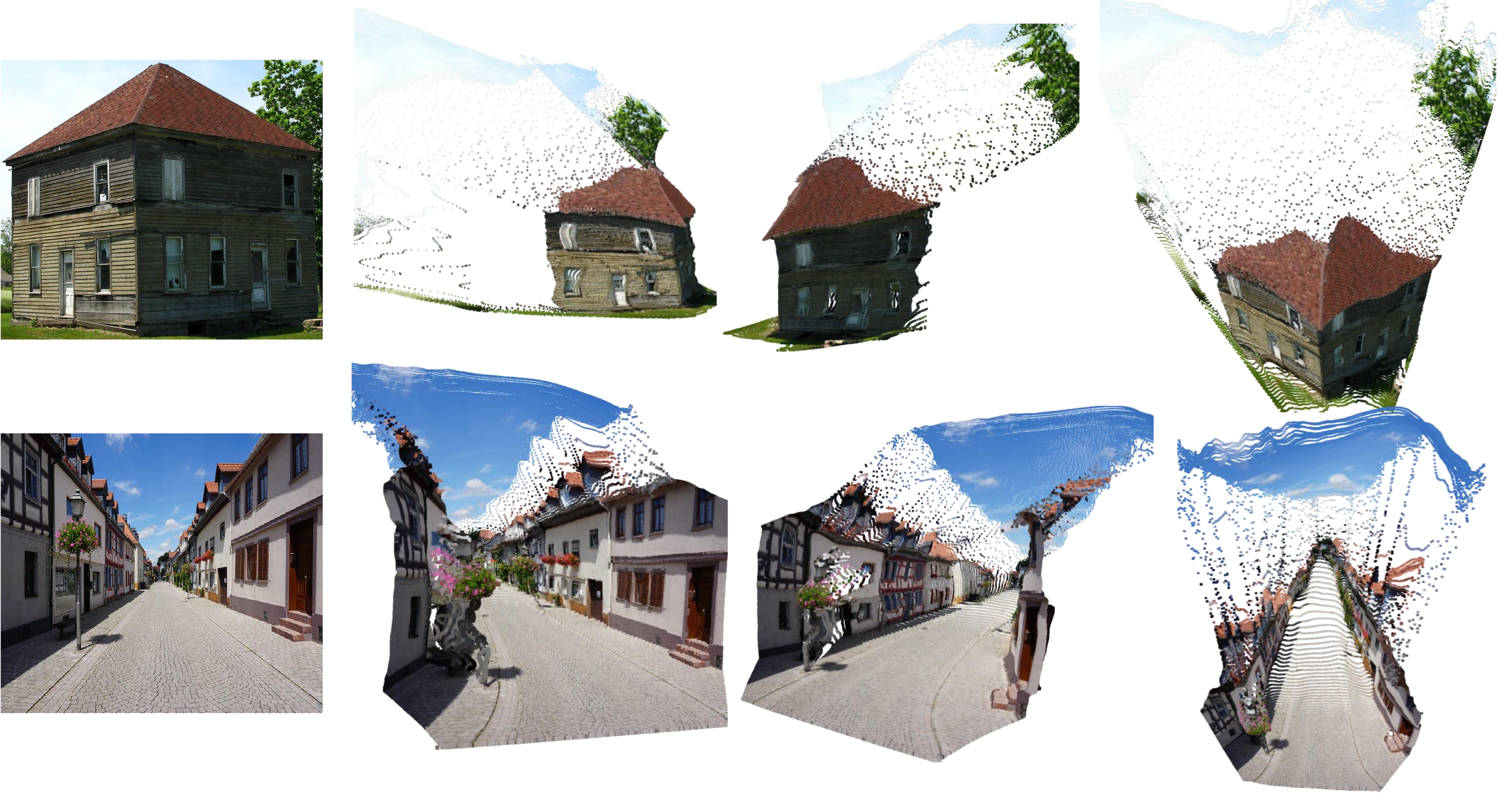}
		\vspace{-1em}
}
{
	\includegraphics[width=0.95\linewidth]{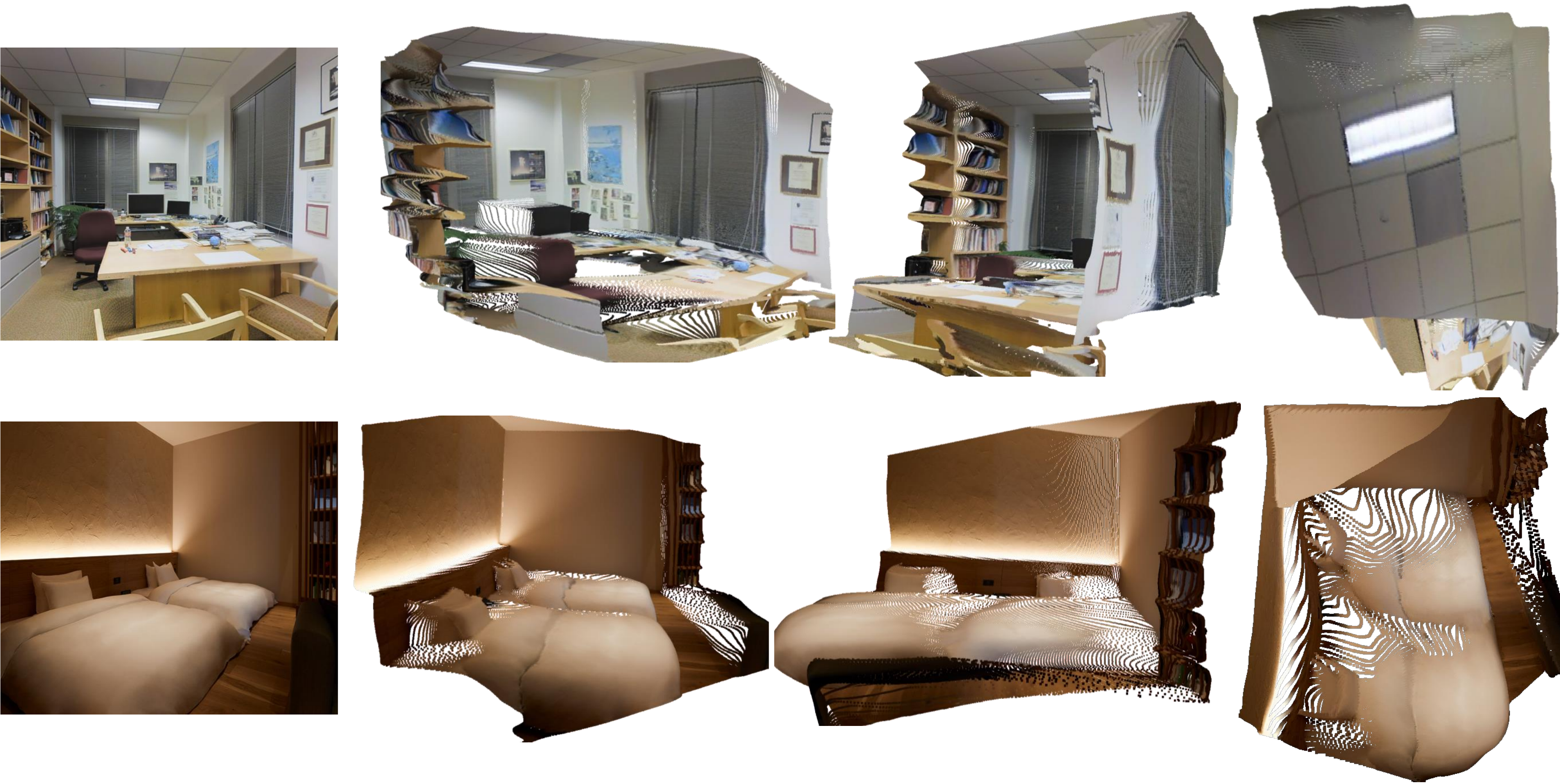}
}
{
	\includegraphics[width=0.95\linewidth]{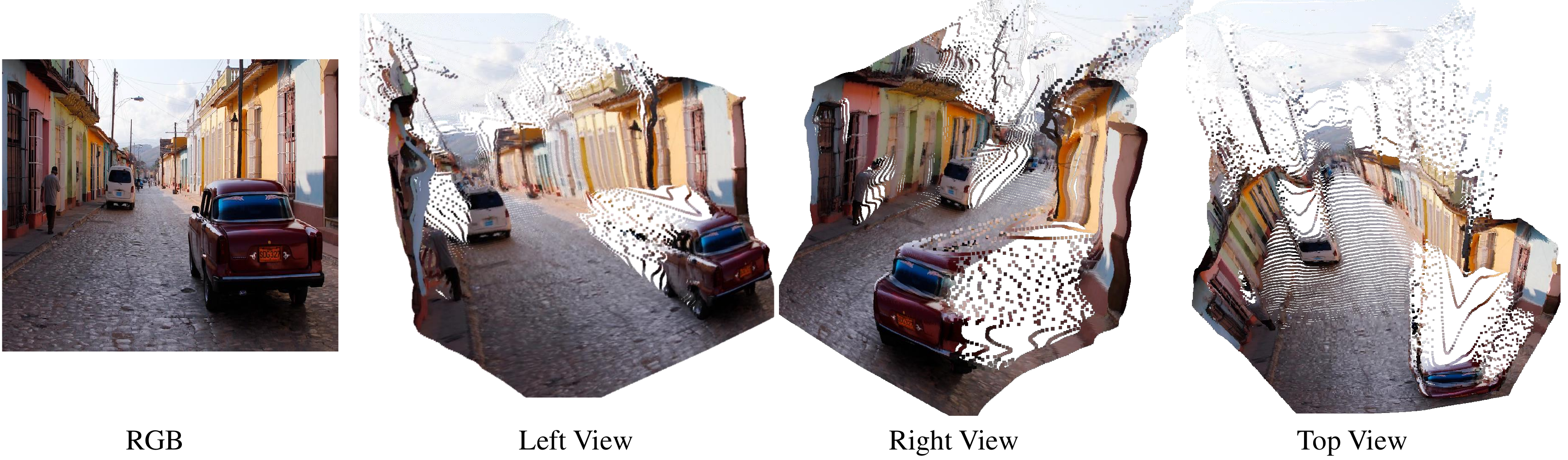}
}
\caption{\textbf{Point Cloud Illustration.} The first column shows the input images. The remaining columns show the point cloud recovered from our proposed approach from the left, right, and top respectively.} %
\label{Fig: pcd_show2}  %
\end{figure*}

We illustrate some examples of the reconstructed 3D point cloud from our proposed approach in Fig.~\ref{Fig: pcd_show2}. All these data are unseen during training. This shows that our method demonstrates good generalizability on in-the-wild scenes and can recover realistic shape of a wide range of scenes.

\begin{figure*}
{
	\includegraphics[width=1\linewidth]{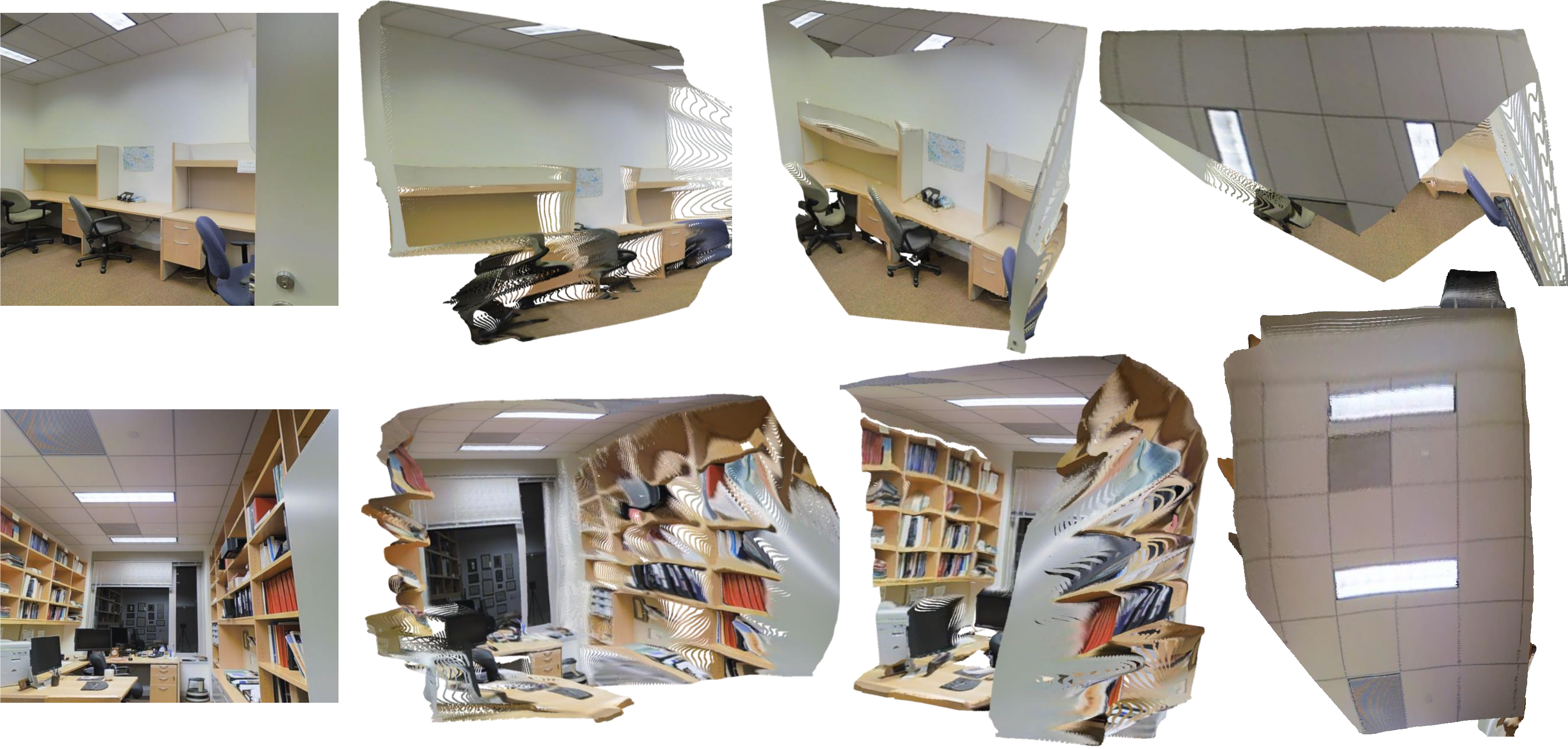}
		\vspace{-1em}
}
{
	\includegraphics[width=1\linewidth]{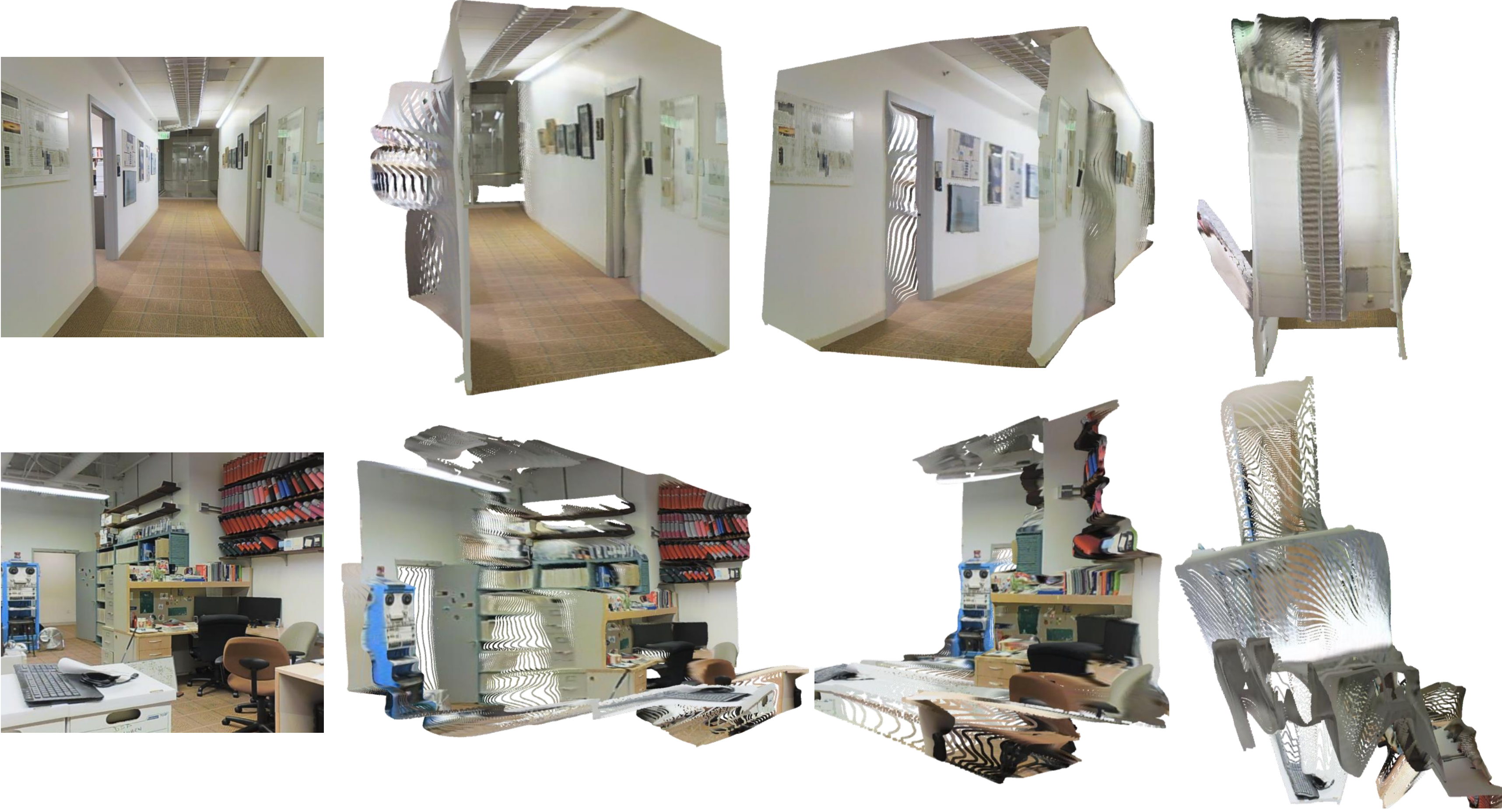}
}
{
	\includegraphics[width=1\linewidth]{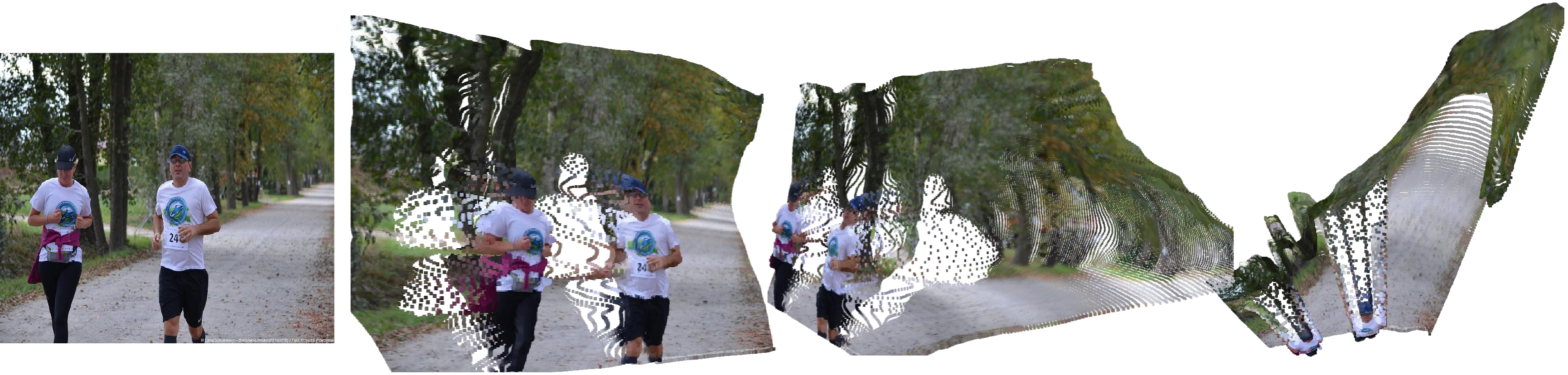}
}
\\
\label{Fig: pcd_show1}  %
\end{figure*}

\section{Illustration of Depth Prediction In the Wild}
\begin{figure*}
{
	\includegraphics[width=1\textwidth]{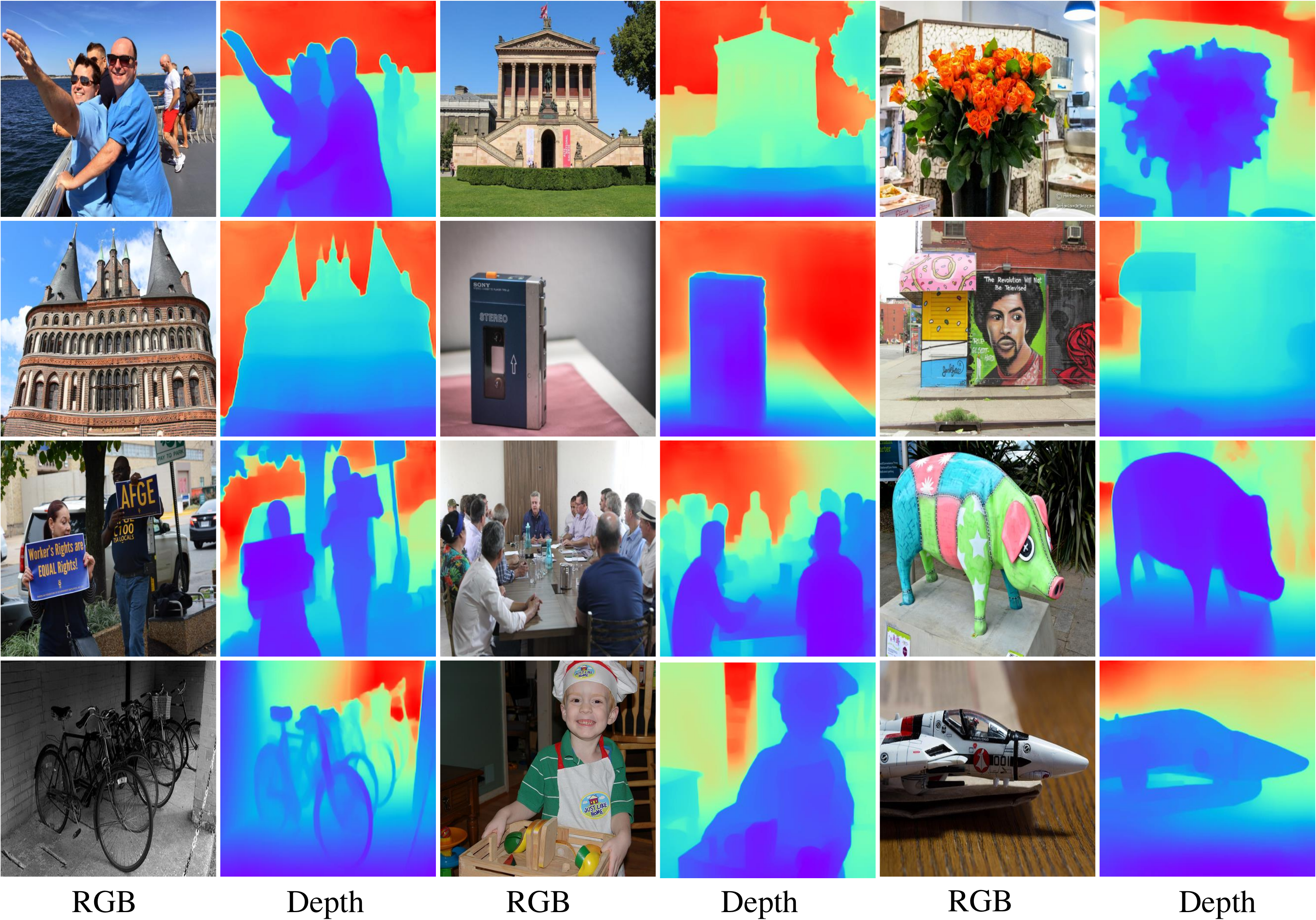}
}
\caption{Examples of depths on in-the-wild scenes. Purple indicates closer regions whereas red indicates farther regions.} %
\label{Fig: depth_cmp4}  %
\end{figure*}

We illustrate examples of our single image depth prediction results in Fig.~\ref{Fig: depth_cmp4}. The images are randomly sampled from DIW and OASIS, which are unseen during training.  On these diverse scenes, our method predicts reasonably accurate depth maps, in terms of global structure and local details.

\begin{figure*}
\centering    %
{
	\includegraphics[width=0.97\linewidth]{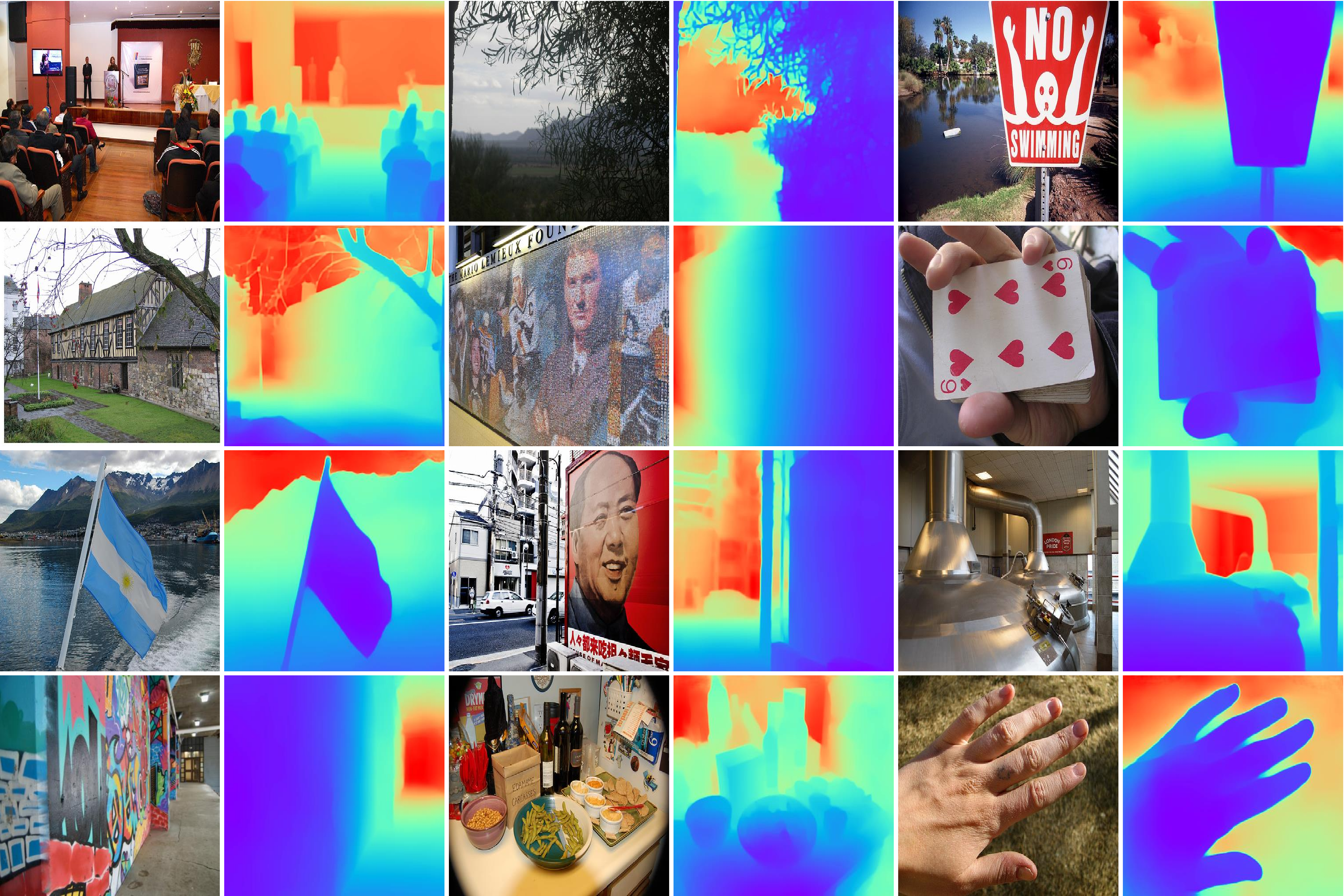}
	\vspace{-1em}
}
{
	\includegraphics[width=0.97\linewidth]{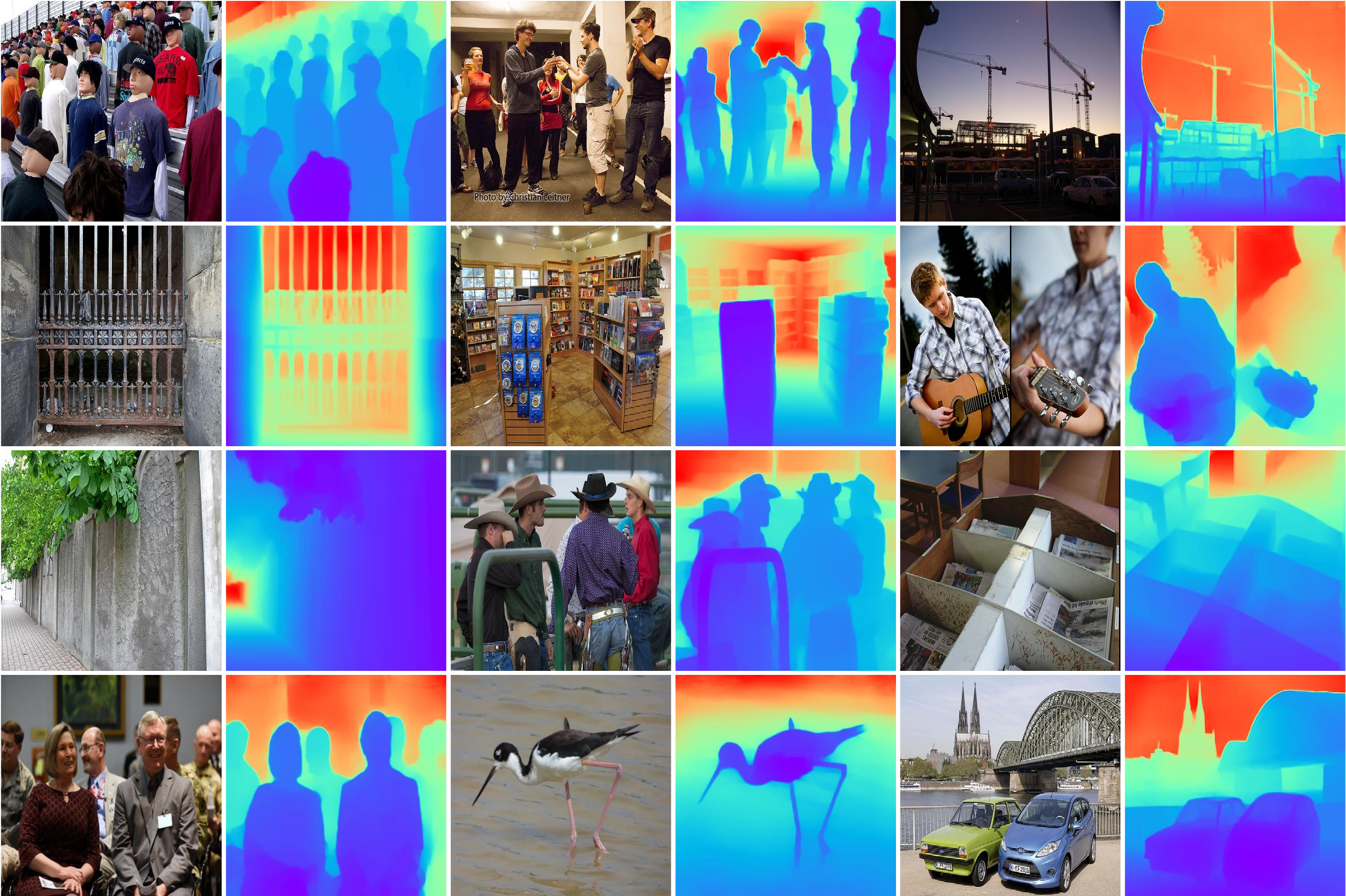}
}
\label{Fig: depth_cmp1}  %
\end{figure*}

\begin{figure*}
\centering    %
{
	\includegraphics[width=0.97\textwidth]{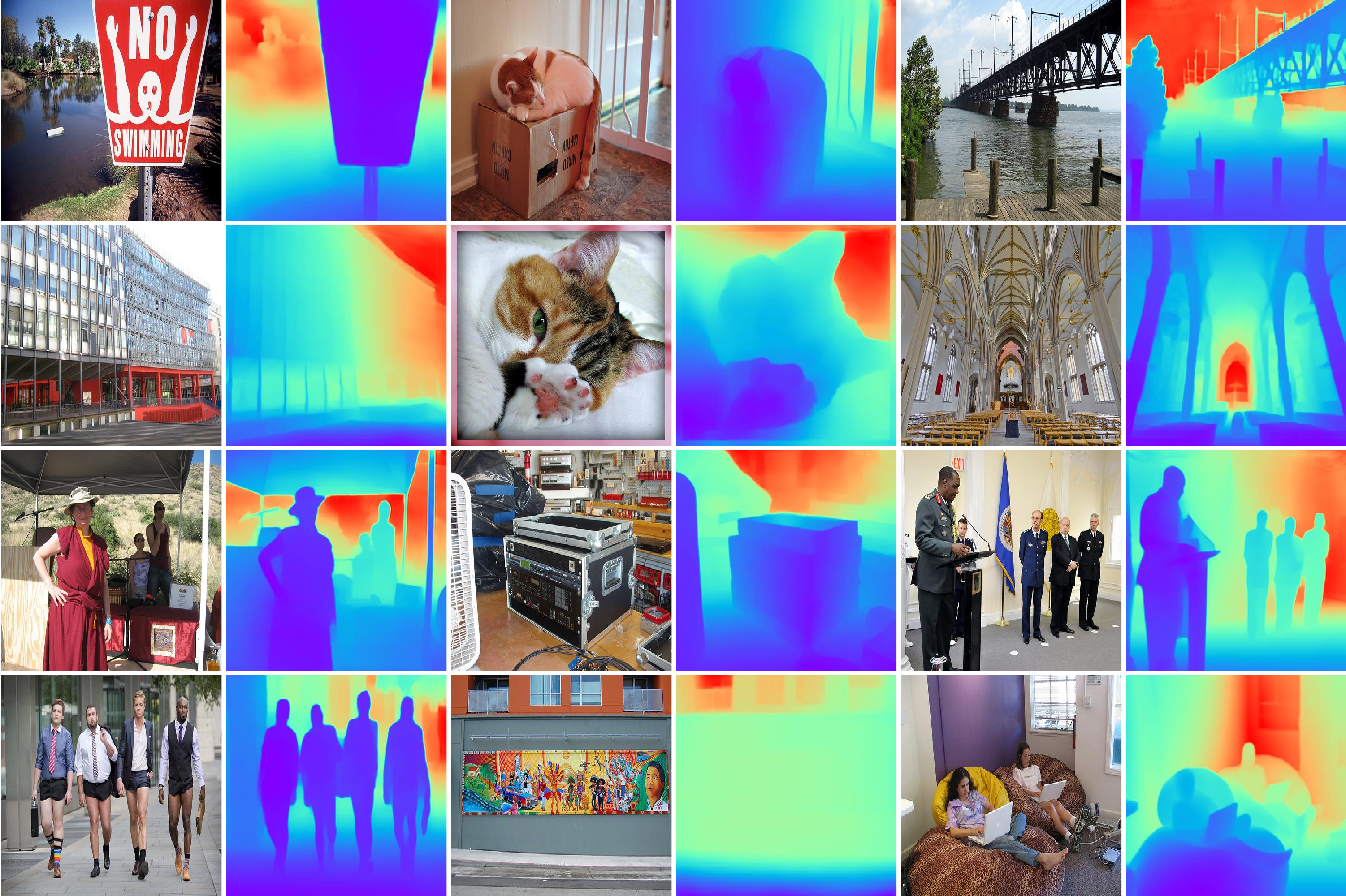}
		\vspace{-1em}
}

{
	\includegraphics[width=0.97\textwidth]{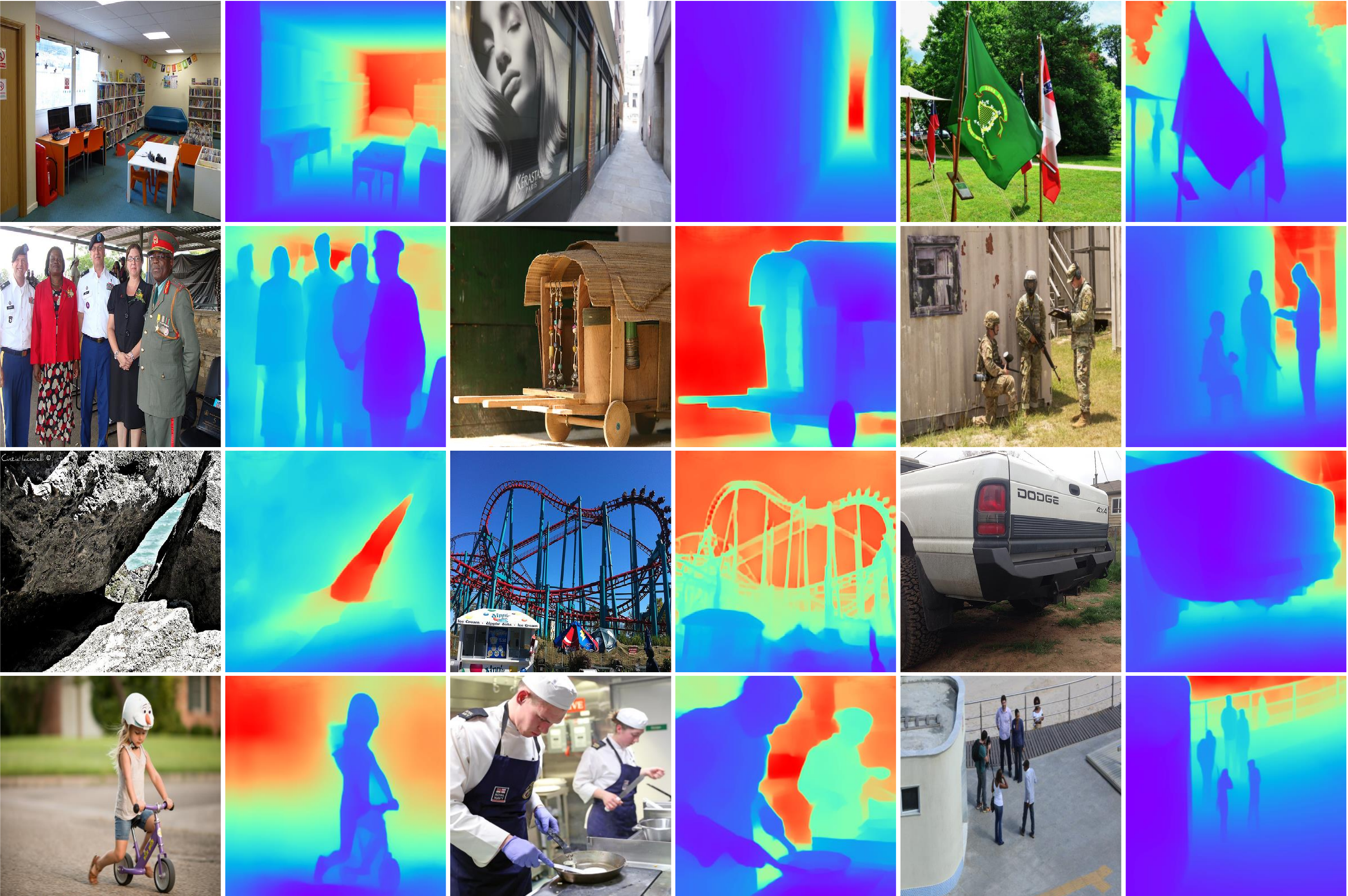}
}
\\
\label{Fig: depth_cmp2}  %
\end{figure*}

\begin{figure*}
{
	\includegraphics[width=0.97\textwidth]{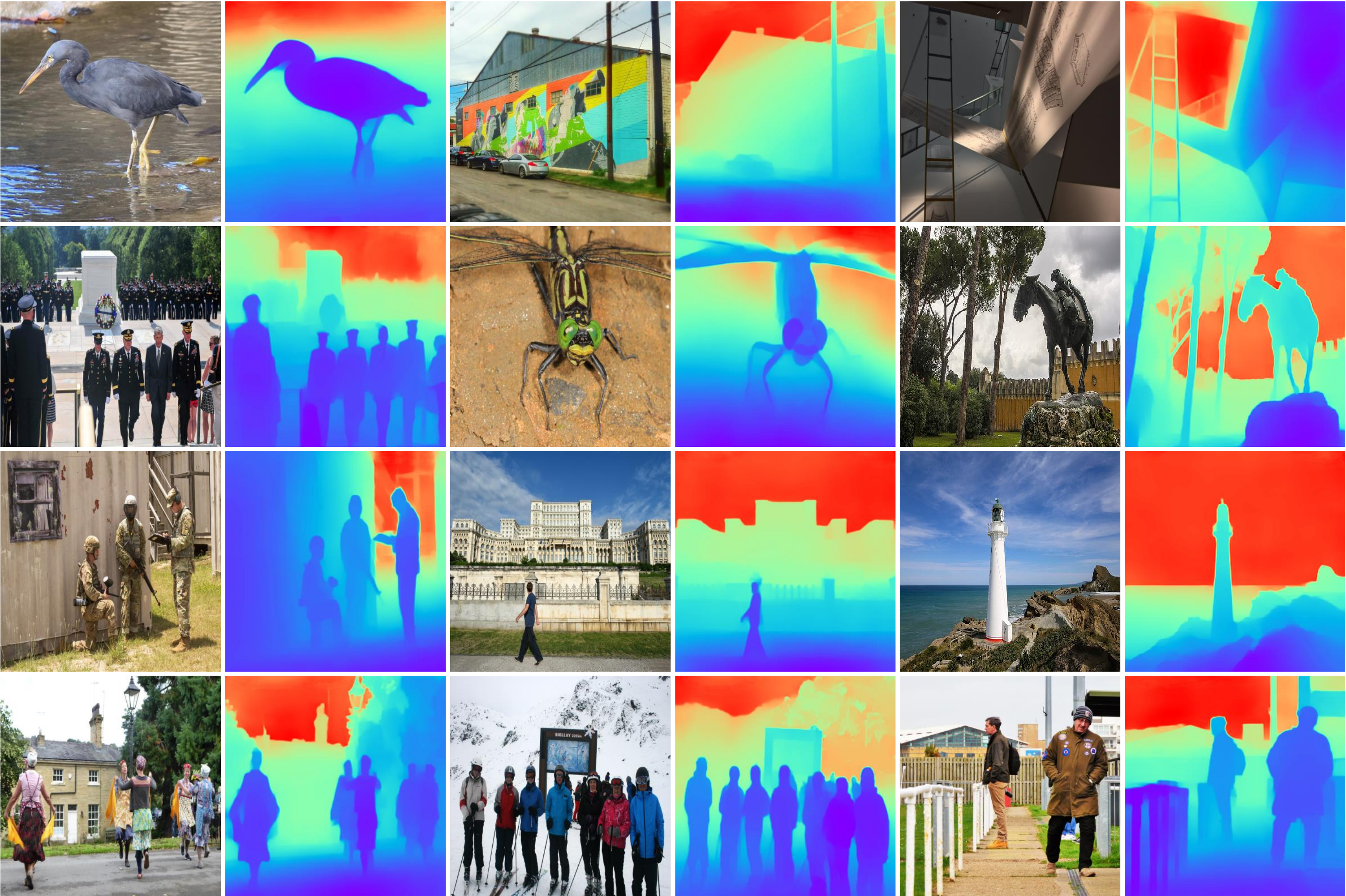}
		\vspace{-1em}
}
{
	\includegraphics[width=0.97\textwidth]{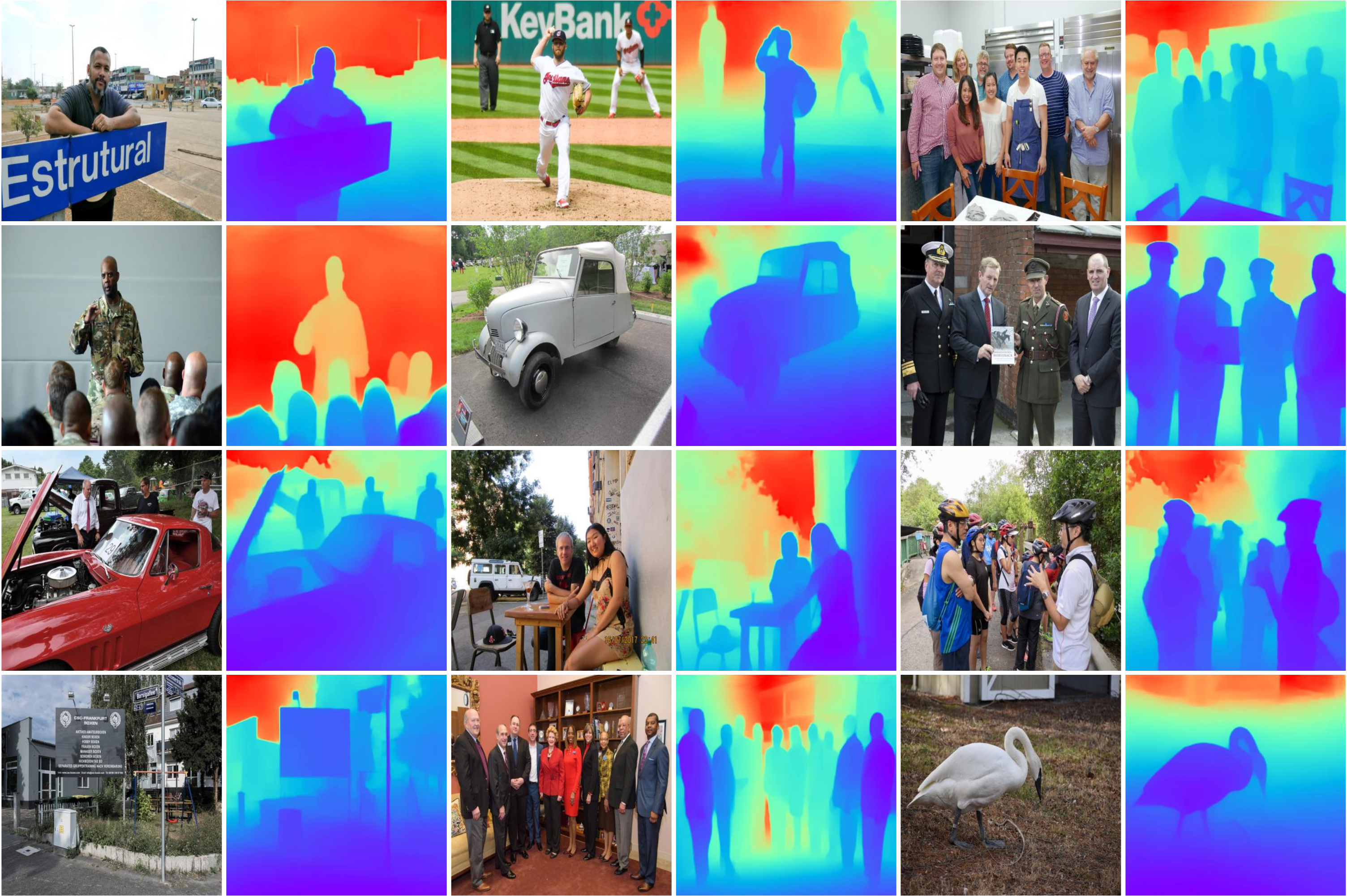}
}\\
\label{Fig: depth_cmp3}  %
\end{figure*}

\section*{Acknowledgment}
This work was in part supported by ARC DP Project ``Deep learning that scales''.

{\small
\bibliographystyle{ieee_fullname}
\bibliography{draft}

\begin{thebibliography}{10}\itemsep=-1pt

\bibitem{armeni2017joint}
Iro Armeni, Sasha Sax, Amir~R Zamir, and Silvio Savarese.
\newblock Joint 2d-3d-semantic data for indoor scene understanding.
\newblock {\em arXiv preprint arXiv:1702.01105}, 2017.

\bibitem{barron2014shape}
Jonathan~T Barron and Jitendra Malik.
\newblock Shape, illumination, and reflectance from shading.
\newblock {\em {IEEE} Trans. Pattern Anal. Mach. Intell.}, 37(8):1670--1687,
  2014.

\bibitem{bian2019unsupervised}
Jiawang Bian, Zhichao Li, Naiyan Wang, Huangying Zhan, Chunhua Shen, Ming-Ming
  Cheng, and Ian Reid.
\newblock Unsupervised scale-consistent depth and ego-motion learning from
  monocular video.
\newblock In {\em Proc. Advances in Neural Inf. Process. Syst.}, pages 35--45,
  2019.

\bibitem{Butler:ECCV:2012}
D.~J. Butler, J. Wulff, G.~B. Stanley, and M.~J. Black.
\newblock A naturalistic open source movie for optical flow evaluation.
\newblock In {\em Proc. Eur. Conf. Comp. Vis.}, pages 611--625. Springer, 2012.

\bibitem{deeplabv3plus2018}
Liang-Chieh Chen, Yukun Zhu, George Papandreou, Florian Schroff, and Hartwig
  Adam.
\newblock Encoder-decoder with atrous separable convolution for semantic image
  segmentation.
\newblock In {\em Proc. Eur. Conf. Comp. Vis.}, 2018.

\bibitem{chen2016single}
Weifeng Chen, Zhao Fu, Dawei Yang, and Jia Deng.
\newblock Single-image depth perception in the wild.
\newblock In {\em Proc. Advances in Neural Inf. Process. Syst.}, pages
  730--738, 2016.

\bibitem{chen2019learning}
Weifeng Chen, Shengyi Qian, and Jia Deng.
\newblock Learning single-image depth from videos using quality assessment
  networks.
\newblock In {\em Proc. IEEE Conf. Comp. Vis. Patt. Recogn.}, pages 5604--5613,
  2019.

\bibitem{chen2020oasis}
Weifeng Chen, Shengyi Qian, David Fan, Noriyuki Kojima, Max Hamilton, and Jia
  Deng.
\newblock Oasis: A large-scale dataset for single image 3d in the wild.
\newblock In {\em Proc. IEEE Conf. Comp. Vis. Patt. Recogn.}, pages 679--688,
  2020.

\bibitem{dai2017scannet}
Angela Dai, Angel~X Chang, Manolis Savva, Maciej Halber, Thomas Funkhouser, and
  Matthias Nie{\ss}ner.
\newblock Scannet: Richly-annotated 3d reconstructions of indoor scenes.
\newblock In {\em Proc. IEEE Conf. Comp. Vis. Patt. Recogn.}, pages 5828--5839,
  2017.

\bibitem{deutscher2002automatic}
Jonathan Deutscher, Michael Isard, and John MacCormick.
\newblock Automatic camera calibration from a single manhattan image.
\newblock In {\em Proc. Eur. Conf. Comp. Vis.}, pages 175--188. Springer, 2002.

\bibitem{eigen2014depth}
David Eigen, Christian Puhrsch, and Rob Fergus.
\newblock Depth map prediction from a single image using a multi-scale deep
  network.
\newblock In {\em Proc. Advances in Neural Inf. Process. Syst.}, pages
  2366--2374, 2014.

\bibitem{facil2019cam}
Jose~M Facil, Benjamin Ummenhofer, Huizhong Zhou, Luis Montesano, Thomas Brox,
  and Javier Civera.
\newblock Cam-convs: camera-aware multi-scale convolutions for single-view
  depth.
\newblock In {\em Proc. IEEE Conf. Comp. Vis. Patt. Recogn.}, pages
  11826--11835, 2019.

\bibitem{fukunaga2013introduction}
Keinosuke Fukunaga.
\newblock {\em Introduction to Statistical Pattern Recognition}.
\newblock Elsevier, 2013.

\bibitem{garcia2015data}
Salvador Garc{\'\i}a, Juli{\'a}n Luengo, and Francisco Herrera.
\newblock {\em Data preprocessing in data mining}.
\newblock Springer, 2015.

\bibitem{geiger2012we}
Andreas Geiger, Philip Lenz, and Raquel Urtasun.
\newblock Are we ready for autonomous driving? the kitti vision benchmark
  suite.
\newblock In {\em Proc. IEEE Conf. Comp. Vis. Patt. Recogn.}, pages 3354--3361.
  IEEE, 2012.

\bibitem{monodepth2}
Cl{\'{e}}ment Godard, Oisin {Mac Aodha}, Michael Firman, and Gabriel~J.
  Brostow.
\newblock Digging into self-supervised monocular depth prediction.
\newblock In {\em Proc. IEEE Int. Conf. Comp. Vis.}, 2019.

\bibitem{hartley2003multiple}
Richard Hartley and Andrew Zisserman.
\newblock {\em Multiple view geometry in computer vision}.
\newblock Cambridge university press, 2003.

\bibitem{he2016deep}
Kaiming He, Xiangyu Zhang, Shaoqing Ren, and Jian Sun.
\newblock Deep residual learning for image recognition.
\newblock In {\em Proc. IEEE Conf. Comp. Vis. Patt. Recogn.}, pages 770--778,
  2016.

\bibitem{hold2018perceptual}
Yannick Hold-Geoffroy, Kalyan Sunkavalli, Jonathan Eisenmann, Matthew Fisher,
  Emiliano Gambaretto, Sunil Hadap, and Jean-Fran{\c{c}}ois Lalonde.
\newblock A perceptual measure for deep single image camera calibration.
\newblock In {\em Proc. IEEE Conf. Comp. Vis. Patt. Recogn.}, pages 2354--2363,
  2018.

\bibitem{hua2020holopix50k}
Yiwen Hua, Puneet Kohli, Pritish Uplavikar, Anand Ravi, Saravana Gunaseelan,
  Jason Orozco, and Edward Li.
\newblock Holopix50k: A large-scale in-the-wild stereo image dataset.
\newblock In {\em IEEE Conf. Comput. Vis. Pattern Recog. Worksh.}, June 2020.

\bibitem{IMKDB17}
E. Ilg, N. Mayer, T. Saikia, M. Keuper, A. Dosovitskiy, and T. Brox.
\newblock Flownet 2.0: Evolution of optical flow estimation with deep networks.
\newblock In {\em Proc. IEEE Conf. Comp. Vis. Patt. Recogn.}, 2017.

\bibitem{karpenko2006smoothsketch}
Olga~A Karpenko and John Hughes.
\newblock Smoothsketch: 3d free-form shapes from complex sketches.
\newblock In {\em ACM. T. Graph. (SIGGRAPH)}, pages 589--598. 2006.

\bibitem{kim2018deep}
Youngjung Kim, Hyungjoo Jung, Dongbo Min, and Kwanghoon Sohn.
\newblock Deep monocular depth estimation via integration of global and local
  predictions.
\newblock {\em {IEEE} Trans. Image Process.}, 27(8):4131--4144, 2018.

\bibitem{Koch18:ECS}
Tobias Koch, Lukas Liebel, Friedrich Fraundorfer, and Marco K{\"o}rner.
\newblock Evaluation of {CNN}-based single-image depth estimation methods.
\newblock In {\em Eur. Conf. Comput. Vis. Worksh.}, pages 331--348, 2018.

\bibitem{lang2010nonlinear}
Manuel Lang, Alexander Hornung, Oliver Wang, Steven Poulakos, Aljoscha Smolic,
  and Markus Gross.
\newblock Nonlinear disparity mapping for stereoscopic 3d.
\newblock {\em {ACM} Trans. Graph.}, 29(4):1--10, 2010.

\bibitem{li2018megadepth}
Zhengqi Li and Noah Snavely.
\newblock Megadepth: Learning single-view depth prediction from internet
  photos.
\newblock In {\em Proc. IEEE Conf. Comp. Vis. Patt. Recogn.}, pages 2041--2050,
  2018.

\bibitem{liu2015learning}
Fayao Liu, Chunhua Shen, Guosheng Lin, and Ian Reid.
\newblock Learning depth from single monocular images using deep convolutional
  neural fields.
\newblock {\em {IEEE} Trans. Pattern Anal. Mach. Intell.}, 38(10):2024--2039,
  2015.

\bibitem{liu2019training}
Yifan Liu, Bohan Zhuang, Chunhua Shen, Hao Chen, and Wei Yin.
\newblock Training compact neural networks via auxiliary overparameterization.
\newblock {\em arXiv preprint arXiv:1909.02214}, 2019.

\bibitem{liu2019pvcnn}
Zhijian Liu, Haotian Tang, Yujun Lin, and Song Han.
\newblock Point-voxel cnn for efficient 3d deep learning.
\newblock In {\em Proc. Advances in Neural Inf. Process. Syst.}, 2019.

\bibitem{Niklaus_TOG_2019}
Simon Niklaus, Long Mai, Jimei Yang, and Feng Liu.
\newblock 3d ken burns effect from a single image.
\newblock {\em {ACM} Trans. Graph.}, 38(6):184:1--184:15, 2019.

\bibitem{prados2005shape}
Emmanuel Prados and Olivier Faugeras.
\newblock Shape from shading: a well-posed problem?
\newblock In {\em Proc. IEEE Conf. Comp. Vis. Patt. Recogn.}, volume~2, pages
  870--877. IEEE, 2005.

\bibitem{Ranftl2020}
Ren\'{e} Ranftl, Katrin Lasinger, David Hafner, Konrad Schindler, and Vladlen
  Koltun.
\newblock Towards robust monocular depth estimation: Mixing datasets for
  zero-shot cross-dataset transfer.
\newblock {\em {IEEE} Trans. Pattern Anal. Mach. Intell.}, 2020.

\bibitem{saito2019pifu}
Shunsuke Saito, Zeng Huang, Ryota Natsume, Shigeo Morishima, Angjoo Kanazawa,
  and Hao Li.
\newblock Pifu: Pixel-aligned implicit function for high-resolution clothed
  human digitization.
\newblock In {\em Proc. IEEE Conf. Comp. Vis. Patt. Recogn.}, pages 2304--2314,
  2019.

\bibitem{saito2020pifuhd}
Shunsuke Saito, Tomas Simon, Jason Saragih, and Hanbyul Joo.
\newblock Pifuhd: Multi-level pixel-aligned implicit function for
  high-resolution 3d human digitization.
\newblock In {\em Proc. IEEE Conf. Comp. Vis. Patt. Recogn.}, pages 84--93,
  2020.

\bibitem{saxena2008make3d}
Ashutosh Saxena, Min Sun, and Andrew~Y Ng.
\newblock Make3d: Learning 3d scene structure from a single still image.
\newblock {\em {IEEE} Trans. Pattern Anal. Mach. Intell.}, 31(5):824--840,
  2008.

\bibitem{schops2017multi}
Thomas Schops, Johannes~L Schonberger, Silvano Galliani, Torsten Sattler,
  Konrad Schindler, Marc Pollefeys, and Andreas Geiger.
\newblock A multi-view stereo benchmark with high-resolution images and
  multi-camera videos.
\newblock In {\em Proc. IEEE Conf. Comp. Vis. Patt. Recogn.}, pages 3260--3269,
  2017.

\bibitem{silberman2012indoor}
Nathan Silberman, Derek Hoiem, Pushmeet Kohli, and Rob Fergus.
\newblock Indoor segmentation and support inference from rgbd images.
\newblock In {\em Proc. Eur. Conf. Comp. Vis.}, pages 746--760. Springer, 2012.

\bibitem{singh2019investigating}
Dalwinder Singh and Birmohan Singh.
\newblock Investigating the impact of data normalization on classification
  performance.
\newblock {\em Applied Soft Computing}, page 105524, 2019.

\bibitem{vasiljevic2019diode}
Igor Vasiljevic, Nick Kolkin, Shanyi Zhang, Ruotian Luo, Haochen Wang, Falcon~Z
  Dai, Andrea~F Daniele, Mohammadreza Mostajabi, Steven Basart, Matthew~R
  Walter, et~al.
\newblock Diode: A dense indoor and outdoor depth dataset.
\newblock {\em arXiv preprint arXiv:1908.00463}, 2019.

\bibitem{wang2019web}
Chaoyang Wang, Simon Lucey, Federico Perazzi, and Oliver Wang.
\newblock Web stereo video supervision for depth prediction from dynamic
  scenes.
\newblock In {\em Int. Conf. 3D. Vis.}, pages 348--357. IEEE, 2019.

\bibitem{wang2020sdc}
Lijun Wang, Jianming Zhang, Oliver Wang, Zhe Lin, and Huchuan Lu.
\newblock Sdc-depth: Semantic divide-and-conquer network for monocular depth
  estimation.
\newblock In {\em Proc. IEEE Conf. Comp. Vis. Patt. Recogn.}, pages 541--550,
  2020.

\bibitem{wang2018pixel2mesh}
Nanyang Wang, Yinda Zhang, Zhuwen Li, Yanwei Fu, Wei Liu, and Yu-Gang Jiang.
\newblock Pixel2mesh: Generating 3d mesh models from single {RGB} images.
\newblock In {\em Proc. Eur. Conf. Comp. Vis.}, pages 52--67, 2018.

\bibitem{wang2020foresee}
Xinlong Wang, Wei Yin, Tao Kong, Yuning Jiang, Lei Li, and Chunhua Shen.
\newblock Task-aware monocular depth estimation for 3d object detection.
\newblock In {\em Proc. {AAAI} Conf. Artificial Intell.}, 2020.

\bibitem{workman2015deepfocal}
Scott Workman, Connor Greenwell, Menghua Zhai, Ryan Baltenberger, and Nathan
  Jacobs.
\newblock Deepfocal: A method for direct focal length estimation.
\newblock In {\em Proc. IEEE Int. Conf. Image Process.}, pages 1369--1373.
  IEEE, 2015.

\bibitem{wu2018learning}
Jiajun Wu, Chengkai Zhang, Xiuming Zhang, Zhoutong Zhang, William Freeman, and
  Joshua Tenenbaum.
\newblock Learning shape priors for single-view 3d completion and
  reconstruction.
\newblock In {\em Proc. Eur. Conf. Comp. Vis.}, pages 646--662, 2018.

\bibitem{xian2018monocular}
Ke Xian, Chunhua Shen, Zhiguo Cao, Hao Lu, Yang Xiao, Ruibo Li, and Zhenbo Luo.
\newblock Monocular relative depth perception with web stereo data supervision.
\newblock In {\em Proc. IEEE Conf. Comp. Vis. Patt. Recogn.}, pages 311--320,
  2018.

\bibitem{xian2020structure}
Ke Xian, Jianming Zhang, Oliver Wang, Long Mai, Zhe Lin, and Zhiguo Cao.
\newblock Structure-guided ranking loss for single image depth prediction.
\newblock In {\em Proc. IEEE Conf. Comp. Vis. Patt. Recogn.}, pages 611--620,
  2020.

\bibitem{xiao2012recognizing}
Jianxiong Xiao, Krista~A Ehinger, Aude Oliva, and Antonio Torralba.
\newblock Recognizing scene viewpoint using panoramic place representation.
\newblock In {\em Proc. IEEE Conf. Comp. Vis. Patt. Recogn.}, pages 2695--2702.
  IEEE, 2012.

\bibitem{xie2017aggregated}
Saining Xie, Ross Girshick, Piotr Doll{\'a}r, Zhuowen Tu, and Kaiming He.
\newblock Aggregated residual transformations for deep neural networks.
\newblock In {\em Proc. IEEE Conf. Comp. Vis. Patt. Recogn.}, pages 1492--1500,
  2017.

\bibitem{Yin2019enforcing}
Wei Yin, Yifan Liu, Chunhua Shen, and Youliang Yan.
\newblock Enforcing geometric constraints of virtual normal for depth
  prediction.
\newblock In {\em Proc. IEEE Int. Conf. Comp. Vis.}, 2019.

\bibitem{yin2020diversedepth}
Wei Yin, Xinlong Wang, Chunhua Shen, Yifan Liu, Zhi Tian, Songcen Xu, Changming
  Sun, and Dou Renyin.
\newblock Diversedepth: Affine-invariant depth prediction using diverse data.
\newblock {\em arXiv preprint arXiv:2002.00569}, 2020.

\bibitem{zamir2018taskonomy}
Amir Zamir, Alexander Sax, , William Shen, Leonidas Guibas, Jitendra Malik, and
  Silvio Savarese.
\newblock Taskonomy: Disentangling task transfer learning.
\newblock In {\em Proc. IEEE Conf. Comp. Vis. Patt. Recogn.} IEEE, 2018.

\bibitem{Zhang2019GANet}
Feihu Zhang, Victor Prisacariu, Ruigang Yang, and Philip Torr.
\newblock Ga-net: Guided aggregation net for end-to-end stereo matching.
\newblock In {\em Proc. IEEE Conf. Comp. Vis. Patt. Recogn.}, pages 185--194,
  2019.

\bibitem{zhang2000flexible}
Zhengyou Zhang.
\newblock A flexible new technique for camera calibration.
\newblock {\em {IEEE} Trans. Pattern Anal. Mach. Intell.}, 22(11):1330--1334,
  2000.

\end{thebibliography}
}

\end{document}